\documentclass[12pt]{article}
\usepackage{CJKutf8}
\usepackage{graphicx} 
\usepackage[T1]{fontenc}
\usepackage{ipa}
\usepackage[utf8]{inputenc}
\usepackage{amsmath}
\usepackage{longtable}
\usepackage{appendix}
\usepackage{booktabs} 
\usepackage{float}
\usepackage{multicol}
\usepackage{multirow}
\graphicspath{{images/}}
\usepackage{xcolor}
\usepackage[export]{adjustbox}
\usepackage{natbib}
\usepackage{color}
\usepackage{tipa}
\usepackage{geometry}
\usepackage{hyperref}
\usepackage{comment}
\usepackage{authblk}
\usepackage{textcomp}
\usepackage{rotating}

\title{A new kid on the block: Distributional semantics predicts the word-specific tone signatures of monosyllabic words in conversational Taiwan Mandarin speech.\thanks{Funding: This work was supported by the European Research Council under Grant SUBLIMINAL (\#101054902) awarded to R. Harald Baayen}}

\author[a]{Xiaoyun Jin}
\author[b]{Mirjam Ernestus}
\author[c]{R. Harald Baayen}

\affil[a]{Quantitative Linguistics, Eberhard Karls Universität Tübingen, 72074 Tübingen, Germany\\Email: xiaoyun.jin@uni-tuebingen.de}

\affil[b]{Center for Language Studies, Radboud University, 6525 HT Nijmegen, The Netherlands\\ Email: mirjam.ernestus@ru.nl}

\affil[c]{Quantitative Linguistics, Eberhard Karls Universität Tübingen, 72074 Tübingen, Germany\\ Email: harald.baayen@uni-tuebingen.de}

\date{}

\begin{document}

\begin{CJK}{UTF8}{bsmi} 
\begin{CJK}{UTF8}{gbsn} 
\maketitle

\pagestyle{plain}

\newpage
\begin{abstract}

\noindent 
We present a corpus-based investigation of how the pitch contours of monosyllabic words are realized in spontaneous conversational Mandarin, focusing on the effects of words' meanings. We used the generalized additive model to decompose a given observed pitch contour into a set of component pitch contours that are tied to different control variables and semantic predictors. Even when variables such as word duration, gender, speaker identity, tonal context, vowel height, and utterance position are controlled for, the effect of word remains a strong predictor of tonal realization.  We present evidence that this effect of word is a semantic effect: word sense is shown to be a better predictor than word, and heterographic homophones are shown to have different pitch contours. The strongest evidence for the importance of semantics is that the pitch contours of individual word tokens can be predicted from their contextualized embeddings with an accuracy that substantially exceeds a permutation baseline.  For phonetics, distributional semantics is a new kid on the block. Although our findings challenge standard theories of Mandarin tone, they fit well within the theoretical framework of the Discriminative Lexicon Model. \\ \ \\

\noindent
\textbf{Keywords: tone, word-specific tonal realization, conversational speech, monosyllabic words, Taiwan Mandarin,
Discriminative Lexicon Model} 
\end{abstract}

\newpage

\section{Introduction} \label{sec:introduction}

The tone system of standard Mandarin Chinese is described as having four lexical tones (T1: high level; T2: rising; T3: falling-rising; T4: falling) as well as a neutral tone \citep{liuf1924,chao1968grammar}, the realization of which is taken to depend on the preceding tone \citep{chao1930system,chien2021investigating}.  
When monosyllabic Mandarin words are produced carefully in laboratory speech, their tonal contours are similar to the canonical contours, the contours as described in the literature \citep{chao1968grammar,lai2008mandarin} and provided in textbooks and dictionaries. However, the actual contours of tones in connected or spontaneous speech can differ substantially due to tonal co-articulation \citep{shen1990tonal,xu1997contextual,brenner2013acoustics,brenner2015phonetics}. \cite{xu1994production,xu1993contextual} found for Beijing Mandarin that, when the pitch values of neighboring tones at syllable boundaries are similar, such as for the tone sequence T4+T2+T4 (the low tail of the first falling tone fits the starting point of the T2, and the endpoint of this rising tone dovetails well with the starting point for the second falling tone), these tones show the canonical contours, while when the end of a tone does not well match the start of the next tone, the tone contours in running speech may be substantially different.

Not only inter-syllabic co-articulation but also intra-syllabic co-articulation may influence the realization of tonal contours. For instance, the initial consonant of a syllable may shape the syllable's tonal contour\citep{xu2003effects}.  Similarly, various cross-linguistic studies have provided evidence that different vowels may have their own intrinsic pitch properties \citep{ho1976acoustic,ladd1984vowel,bo1987vowel,whalen1995universality}. For instance, \cite{zee1980tone} found that in Taiwan Mandarin, the F0 of the vowel /u/ was the highest, followed by /i/, and then  /a/ and /\textschwa/. The experimental results of \cite{bo1987vowel} suggest, however, that, in standard Mandarin,  the intrinsic pitch of /a/ is not lower than  of /i/ and /u/ when produced with T3 and T4 tones by female speakers. Besides co-articulation, a great deal of studies have shown that prosody \citep{ouyang2015prosody}, especially duration, tends to affect mandarin tones \citep{wang2020interaction}. 

Besides these articulatory constraints, recently, \citet{Chuang:Bell:Tseng:Baayen:2024} reported that the tonal pattern of Taiwan Mandarin disyllabic words with the T2-T4 tone pattern is partially determined by these words' meanings.  This result fits well with previous studies. \citet{plag2015homophony} observed that the duration of English word-final /s/ co-varies with the semantics that this segment is realizing, e.g., plural number on nouns, singular number on verbs, and possession. \citet{gahl_time_2024} showed that the spoken word duration of English homophones is in part predictable from the meanings of these homophones.  The finding of \citet{Chuang:Bell:Tseng:Baayen:2024} has been replicated for disyllabic words with all 20 possible combinations of tones  by \citet{lu2025realization}. In \citet{lu2025realization}, the effect size of words' meaning exceed that of the canonical tone pattern. Considered jointly, these findings support the possibility that also in Mandarin monosyllabic words, the way in which pitch contours are realized is co-determined by semantics.  

Although in the Mandarin lexicon, disyllabic words are more frequent \citep{chen1993some}, in casual speech, monosyllabic words (with 67\%) are more frequent than disyllabic words (with 30\%) \citep{wu2023mandarin}. 
It is at present unclear whether the results obtained by \citet{Chuang:Bell:Tseng:Baayen:2024,lu2025realization} for disyllabic words generalize to monosyllabic words. Compared to disyllabic words, the meanings of monosyllabic words are substantially more ambiguous \citep{chen1992word,lin2010ambiguity}.  For example,  打 (\textit{da3}) can be used in the sense of `to play' as in  打篮球 \textit{da3 lan2-qiu2} `play basketball', in the sense of `hit' as in  打桌子 \textit{da3 zhuo1-zi0} `to hit the table',  and in the sense of `to give a call' as in  打电话 \textit{da3 dian4-hua4}. At the same time, monosyllabic words offer novel opportunities to test the hypothesis that words' meanings co-determine their pitch contours. Among disyllabic Mandarin words, heterographic homophones are scarce, and not frequent enough to be well-attested in small corpora of conversational speech.  In contrast, among monosyllabic words, heterographic homophones, such as  弟 \textit{di4}, `brother', and  地 \textit{di4} `ground',  are more widespread.  If indeed words' meanings co-determine the realization of tone, then the prediction follows that the tonal realizations of heterographic homophones must be somewhat different. 

The goal of the present study is to clarify, for spontaneous spoken Taiwan Mandarin, whether word, and especially words' meanings, co-determine the pitch contours  of monosyllabic words with the five canonical tones \citep[as described in][]{chao1930system,xu1997contextual,peng1997production}, like they do for disyllabic words, when other factors such as word duration or speech rate \citep{cheng2015mechanism,tang2020acoustic}, segmental properties \citep{ho1976acoustic,ladd1984vowel,bo1987vowel,whalen1995universality}, and tonal context \citep{shen1990tonal,xu1997contextual,xu1993contextual} are controlled for.  A related question is whether heterographic homophones, which share their segments and canonical tone, have different tonal contours in spontaneous corpus? A final question of interest is whether the lexical tone pattern categories observed in careful standard Mandarin speech \citep{chao1930system,xu1997contextual} and documented in textbooks and dictionaries can also be well distinguished in spontaneous Taiwan Mandarin conversational speech when tonal context is taken into account.   

The present study investigates the contribution of the word or its meaning to its tonal realization for monosyllabic words containing one of the vowels /a, i, u, \textschwa/, and carrying any of the five tones, in a corpus of spontaneous Taiwan Mandarin. 

The remainder of this paper is structured as follows. Section~\ref{sec:data} introduces the corpus that we used, how we extracted and processed the pitch contours, the covariates that we took into consideration, and the method we used for statistical analysis.  Section~\ref{sec:results} reports the results obtained. The implications of our findings are laid out in the general discussion (section~\ref{sec:gendisc}).

\section{Method}\label{sec:data}

\subsection{Data}
\noindent

The corpus used in the current study is the Taiwan Mandarin spontaneous speech corpus \citep{fon2004preliminary}. Taiwan Mandarin is a variety of Mandarin with influences from southern Min \citep{norman1988chinese}. This corpus has been transcribed at the word level using traditional Chinese characters. We followed the transcriptions in the corpus,
and distinguished between word types on the basis of the characters with which the words are transcribed. In Mandarin, single-character words always correspond to single spoken syllables. The maximal structure of a syllable is CGVN: an onset consonant (C), a pre-nuclear glide (G), a vocalic nucleus (V) and a coda consonant (N), which, according to the phonotactic constraints on the Mandarin syllables, must be a nasal \citep{duanmu2007phonology}. Previous studies have found evidence for the merger of the two nasal codas /n/ and /\ng/, notably in southern dialects and also in Taiwan Mandarin \citep{chiu2019uncovering,chiu2021articulatory}. Further, final nasals are often not realized, or realized in the form of nasalization of the preceding vowel \citep{yang2010phonetic}. Whether the prevocalic glide is actually part of a diphthong is under debate \citep{duanmu2007phonology}.  In Mandarin, the vowel can be one of seven monophthongs as well as one of eight diphthongs \citep{yi1920lectures}. We restrict ourselves to the vowels /a, i, u, \textschwa/ and to syllables without coda consonants and glides to be able to better control for segmental effects on the realization of the tones. In this study, we define a \textbf{word token} as the pairing of an acoustic form in the corpus with its corresponding specific meaning in context. \textbf{Word type} (or \textbf{word} for short) is defined as a set of \textbf{word tokens} that share highly similar forms and highly similar  meanings. 
Thus, all sound tokens of \begin{CJK}{UTF8}{bsmi} 媽 \end{CJK} in the sense of `mother' and their corresponding context-specific meanings are regarded as tokens of the same type.

In this corpus, there are 118,592 single-character word tokens, which represent 1471 orthographic word types. 
\footnote{
We follow the transcriptions that come with the corpus, which, as in standard practice, represent  potential phrasal units such as 打折 \textit{da3zhe3} `give a discount' as lexical units on a par with compounds such as 双打 \textit{shuang1da3} `doubles (as in tennis)'.  As a consequence, word tokens that are part of phrasal lexical units and compounds are not included in our dataset. 
}
Among them, there are many word types and tokens of with or without a consonant followed by /a, i, u, \textschwa/: 53,139 tokens of 699 word types. About 85\% of these tokens belong to just 32 high frequency words, such as  的 \textit{de0, `genitive'},  你 \textit{ni3} `you', and  他 \textit{ta1} `he'.  In order to prevent model predictions from being heavily biased by the highest frequency words, we randomly sampled 220 tokens 
as uniformly as possible across the 55 speakers, for words with a token frequency greater than 220. On average, there are four tokens of a word per speaker (to a total of $ 4\times 55 = 220 $ tokens). 
As a further measure to ensure that for statistical analysis there are sufficient tokens for a word type, we excluded those words with a token frequency lower than 10, which left us with 8187 tokens. 

We used the Montreal Forced Aligner (henceforth MFA) \citep{mcauliffe2017montreal} to determine the boundary between the consonant and vowel in the words under study. Among the 8187 tokens, 408 tokens were not assigned a segmental boundary, typically due to being too short. Forced-aligned text grids were then audited manually in Praat \citep{boersma19922022} to ensure that segment boundaries were correctly placed. Due to unclear vowel pronunciations, potentially caused by vowel reduction or background noise, 304 tokens were excluded during manual verification. Subsequently, F0 for the vowel of target tokens was measured using the method described in \cite{wempe2018sound}, implemented in a script for Praat. The script uses auto-correlation (without filtering) to determine pitch, with the pitch floor set at 75 Hz and the pitch ceiling at 500 Hz. Every 5 ms, the F0 value was extracted. Within word octave jumps were only found in fewer than about 0.08\% of the F0 data points, typically only effecting one measurement point for any given token. We excluded these data points while retaining the remaining data points of these tokens. 

Unsurprisingly, for longer vowels, more pitch measurements were available. On average, each token had about 19 measurement points. However, 12\% of the tokens had fewer than 5 measurement points, which is problematic for statistical modeling of the pitch contours. We therefore removed these tokens from our dataset.  The resulting dataset comprised 6555 tokens of 95 word types.

The meaning of every word token in the dataset was tagged using a word sense disambiguation system \citep{hsieh2024resolvingregularpolysemynamed} based on the Chinese WordNet \citep{huang2010constructing}. For the 95 word types in our dataset, a total of 406 word senses was identified. Most of the word types have one to three word senses. 70\% of the word types have more than one word sense. Because word sense is one of the main predictors in this study,  we ensured that all word senses had sufficient tokens for model fitting. We therefore included, for a given word type, only those tokens whose word senses were represented by at least another 3 tokens. This left us with a dataset of 6120 tokens, representing 196 word senses across 95 word types.  

Every speaker produced multiple word types, and every word was produced by multiple speakers. The median number of word types per speaker is 37, the mean is about 37 and the range is 28 to 48. Likewise, every word type was pronounced by multiple speakers, although there is quite some variation in the number of speakers per word type (e.g. 土 is produced by two speakers only, while, at the other extreme, 他 is produced by all 55 speakers). On average, every speaker contributed 21 words (median 16) to the dataset.  Table~\ref{tab:vc} provides an overview of the distributions of word-initial consonants for the four vowels under investigation.

\begin{table}[htbp]
\centering

\large
\begin{adjustbox}{max width=\textwidth}
{\bfseries
\begin{tabular}{|c|c|c|c|c|c|c|c|c|c|}
\hline
vowel & nasal & \multicolumn{2}{c|}{plosive} & \multicolumn{2}{c|}{affricate} & fricative & lateral&  NULL&N \\
\cline{2-10}
   & & unaspirated & aspirated & unaspirated & aspirated  & & & &\\\hline
a & /m/(481); /n/(411) & /p/(355); /t/(345) & /p\super h/(40); /t\super h/(667)   &  /t\textrtails/(8)& /ts\super h/(6) ; /t\textrtails\super h/(92)&/f/(6);  /\textrtails/(33)& /l/(132) &(158) &2734\\ \hline

\textschwa & /m/(8); /n/(209) & /t/(369);/k/(266) & /k\super h/(113)  &  /t\textrtails/(226)& /t\textrtails\super h/(34); /\textrtails/(6)&/\textctz/(16);/h/(38) & /l/(115)&(8) &1408\\ \hline

u & & /p/(190);/t/(97) & /k\super h/(28);/t\super h/(13)&/ts(33);t\textrtails/(222)  & t\textrtails\super h/(53) &/f/(11); \textrtails/(66);/\textctz/(8)& /l/(41)& &762\\ \hline

i & /n/(246)& /p/(86);/t/(63) & t\super h/(34)&/t\textctc(178)  & /t\textctc\super h/(118) & \textctc/(86)& /l/(146)& (259) &1216\\ \hline

\end{tabular}
}
\end{adjustbox}
\caption{Distribution of consonants broken down by vowel.}
\label{tab:vc}

\end{table}

\subsection{Statistical modeling}

\begin{figure}
    \centering
    \includegraphics[width=1\linewidth]{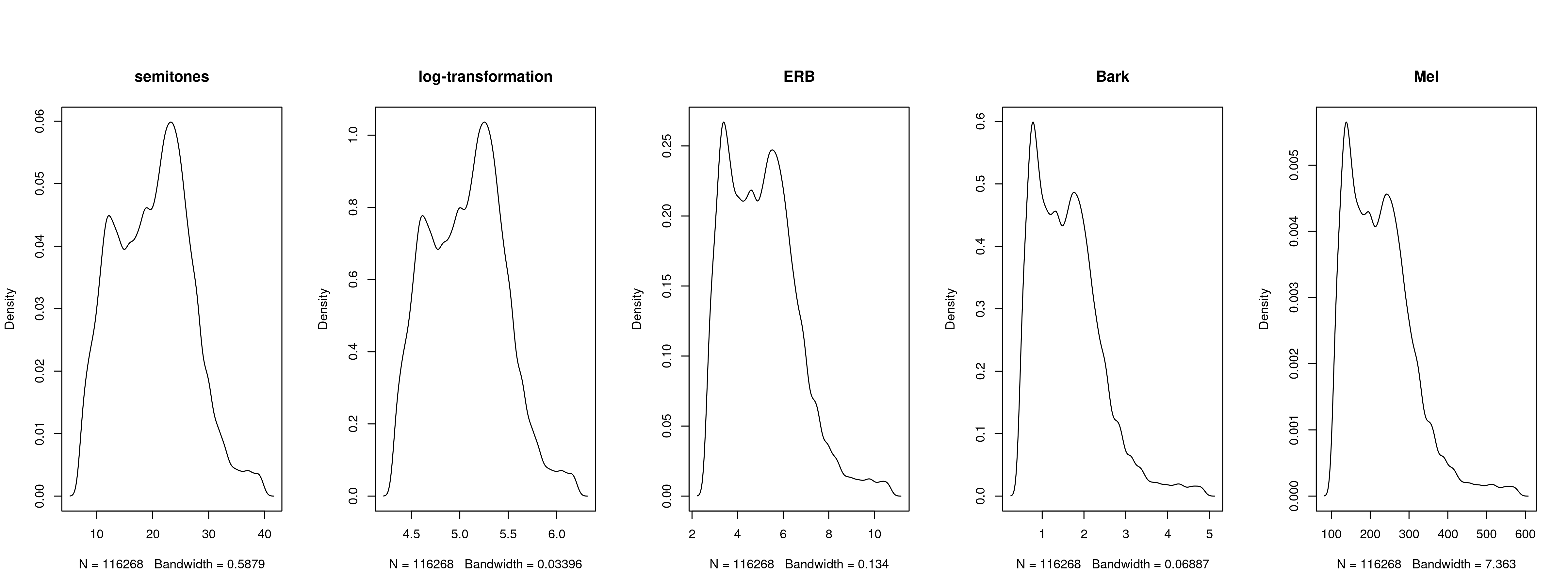}
    \caption{Distribution of F0 values according to five different transformation methods.}
    \label{fig: transformation}
\end{figure}

\noindent
F0 contours are non-linear functions over time. We used the generalized additive model (GAM) to predict tone contours using the \textbf{mgvc} package  \citep{wood2015package} in \citet{r2013r}.  We fitted Gaussian GAMs to the pitch contours, which presupposes that the response variable is roughly normally distributed. As the distribution of pitch measurements has a long right tail, a transformation of the pitch measurements is therefore required. Many different transformations for F0 have been proposed in the literature on auditory perception, such as a transformation to semitones \citep{cole2023enhancement}, the equivalent rectangular bandwidth (ERB transformation) \citep{kitamura2007speaker}, and the transformation to the Bark scale or Mel scale \citep{nolan2003intonational}. As can be seen in Figure~\ref{fig: transformation}, the semitone and the natural logarithmic transformation result in distributions with the least skew. We selected the natural log transformation as our primary focus is on the production of tone rather than on perception.

Pitch contours are time series with slowly changing values that are inevitably auto-correlated.  One common strategy to deal with auto-correlation in GAMs is to incorporate an AR(1) process \citep{Wood:2017,baayen2018autocorrelated}.   Inclusion of an AR(1) process with auto-correlation of about $\rho  = 0.9$ removed nearly all auto-correlation from the residuals.\footnote{The  distribution of the residuals of a Gaussian model with AR(1) has heavy tails which resist correction by assuming the residuals are following a scaled t-distribution \citep[see also][]{Chuang:Bell:Tseng:Baayen:2024}.}  We normalized time to the interval [0,1]. Time normalization is essential for regression modeling of the shape of pitch contours with GAMs. Longer words had more measurement points in the [0,1] interval. 

The statistical models reported below take into consideration the following  two \textbf{speaker-related control variables}.

\begin{description}
\item[Gender] Gender serves as a main control variable, as female speakers tend to speak with a higher pitch and wider pitch ranges than male speakers \citep{gelfer2005relative,shen2011effect}.

\item[Speaker] To allow for differences in the average pitch contours of individual speakers,  we included a by-speaker random smooth in our statistical models. 
\end{description}

In addition, we took the following two \textbf{context-related control variables} into account.
\begin{description}

\item[Tone sequence] Tones show co-articulation with adjacent tones \citep{shen1990tonal,mok2004effects}. To enable the GAM to account for the influence of neighboring tones on the realization of a word's pitch contour, we looked up  the pinyin of the preceding and following syllables, and extracted the canonical tones of these syllables from the pinyin.  There are 36 possible combinations of the preceding and following tone: each possible combination of the four lexical tones, one neutral tone, and NULL when the target word occurs next to a pause. All 36 possible combinations are attested in our dataset. As the effect of adjacent tones is expected to vary depending on the  tone of the syllable under study, we created a factor, \texttt{tone sequence}, with as levels all attested pairings of 36 neighboring tones crossed with the 5 tones attested for our target syllables. Of the 180 possible combinations, 141 were attested in our dataset.  It is important to keep in mind that \texttt{tone sequence} is established on the basis of the canonical tones of the target word and its immediate neighbors, and that \texttt{tone sequence} thus does not take into account that the tones of the immediate neighbors themselves may be subject to co-articulation with their neighboring tones. 

\item[Utterance position] As the realization of F0 contours in utterances is co-determined by sentence intonation \citep{ho1976acoustic,yuan2011perception}, we calculated the 0-1 normalized position of a target word in its utterance as covariate, following \citet{Chuang:Bell:Tseng:Baayen:2024}. It is difficult to differentiate, in spontaneous conversational speech, between utterance boundaries, hesitations, pauses, and interruptions by other speakers. In the present study, an utterance was defined as a sequence of words preceded and followed by a perceivable pause (regardless of its duration), following the annotations provided by the corpus. The normalized position of a given token in an utterance is the position at which the token occurs divided by the total number of words in the utterance. Hence, this predictor is bounded between 0 and 1.

\end{description}

The statistical models incorporated the following \textbf{lexical control variables}.
\begin{description}

\item[Vowel] Various cross-linguistic studies \citep{ho1976acoustic,ladd1984vowel,bo1987vowel,whalen1995universality} have shown that vowels differ in their intrinsic pitch. Specifically, high vowels, such as /i/, tend  to have higher pitch than low vowels. In the current study, we restricted our target tokens to words containing one of the vowels /a,i,u,\textschwa/, which constitute the levels of the factor \texttt{vowel}.

\item[Duration] Previous studies have shown that the realization of pitch contours depends in part on word duration \citep{howie1976acoustical,lin1989contextual,woo1969prosody,yang2017duration}. We included word \texttt{duration} as a predictor.  We  log-transformed duration in order to avoid outlier effects of long word durations. Duration is strongly correlated with speech rate.  Including both \texttt{duration} and speech rate gives rise to high collinearity and concurvity. As \texttt{duration} turned out to be the superior predictor, we included \texttt{duration} as predictor, and do not report further on speech rate.  We included an interaction of duration by gender, and we used a (\texttt{ti()}) tensor product interaction  to allow for an interaction of \texttt{time} and \texttt{duration}.

The three \textbf{core predictors} of central interest to the present study are the following.

\item[Tone pattern]
We included \texttt{tone pattern} as a factorial predictor in our model. The \texttt{tone pattern} is the canonical tone of a word as specified in dictionaries and taught in textbooks \citep{xu1997contextual,peng1997production}. The levels of \texttt{tone pattern} are high (T1), rise (T2), dipping (T3), falling (T4), and neutral (T0) tone. Although in spontaneous speech, contextual factors such as tonal context, and utterance position may alter the shape of the F0 contour of a monosyllabic word, it is generally assumed that these factors do not give rise to qualitative differences in the realization of the canonical tone patterns \citep{xu1997contextual,peng1997production}. In other words, the tonal contours in spontaneous speech are expected to resemble the canonical tone patterns to a considerable extent.  We therefore expect to find an independent contribution of \texttt{tone pattern} to words' pitch contours. 

\item[Word]
The factorial predictor word has as levels the characters of the individual words with which they are represented in the orthographic transcription of the Taiwan Mandarin corpus. Our dataset contains 12 sets of heterographic homophones: nine doublets (\textit{bu4}, \textit{ji3},\textit{ji4}, \textit{ke1}, \textit{li3},\textit{ma0}, \textit{ni3}, \textit{qi2},       \textit{xi4}), two triplets (\textit{de0}, \textit{di4} ), and a quadruplet (\textit{ta1}).  The only character in our dataset that has more than one possible segmental realization is 地, which can be pronounced either as \textit{d{e}0} or \textit{d{i}4}. In addition, a few characters have more than one possible tonal realization, and hence more than one meaning (e.g.,  哪 \textit{na0}/\textit{na3} `exclamatory tone/where'). Such characters are included as different words in our analyses.

\item[Word sense]\label{sec:word sense}
Monosyllabic words in Mandarin often have many different meanings or senses \citep{chung2006mandarin}. In our dataset of 6120 tokens across 95 word types, \texttt{Word sense} is a factorial predictor with 196 levels.  For instance, the character 住 is assigned two word senses: \begin{CJK}{UTF8}{bsmi} 在固定的特定地方或區域進行與居住相關的一切活動 `live at someplace' and 形容前述動作停頓或靜止 `stop' .   
\end{CJK}
\end{description}

In our analyses, we do not include the initial consonant as a lexical control variable, the reason being that in combination with the vowel, statistical models become overspecified. The combination of the vowel and  the tone mostly fully determines the identity of a word. As a consequence, the factor \texttt{word} represents both words' segmental properties as well as words' overall meanings.  

We started out with a baseline model with all control predictors. To this baseline model, we added one additional core predictor, denoted by $X$, which was either \texttt{tone pattern}, \texttt{word}, or \texttt{word sense}. These three representations of $X$ were not simultaneously entered in the model specification because they are too collinear. To assess the relative importance of these three variables, we  used Akaike's information criterion \citep[AIC,][]{Akaike:1974}.  The AIC can be used to compare non-nested models, and provides a measure of the relative quality of these models. Lower values of AIC indicate higher quality. 

Each model had the following specification:

\begin{tabbing}
mm \= \texttt{pitch(logF0)} \= $\sim$ \= \texttt{gender} + \kill
    \> \texttt{pitch(logF0)} \> $\sim$ \> \texttt{gender} +  \texttt{vowel} +\\
    \> \> s(\texttt{normalized time}, by = gender) + \\
    \> \> s(\texttt{duration}, by = gender) + \\
    \> \> ti(\texttt{normalized time}, \texttt{duration}) + \\
    \> \> s(\texttt{normalized time}, \texttt{speaker}, bs = ``fs", m=1) + \\
    \> \> s(\texttt{tone sequence}, bs = ``re") + \\
    \> \> s(\texttt{utterance position}, k = 4) + \\
    \> \> s(\texttt{normalized time}, \texttt{X}, bs = ``fs", m = 1) \\
\end{tabbing}

The first eight terms of this model specification control for \texttt{gender}, \texttt{tone pattern}, \texttt{vowel}, \texttt{normalized time}, log-transformed \texttt{duration}, \texttt{speaker}, \texttt{tone sequence} and \texttt{utterance position}.  The factors \texttt{gender}, \texttt{tone pattern} and \texttt{vowel} are treated as fixed-effect predictors. We considered a three-way interaction between \texttt{gender}, \texttt{normalized time}, and log-transformed \texttt{duration}. 

We fitted a nonlinear random effect (`wiggly curve') for  \texttt{normalized time}  for each individual speaker, using factor smooths.  These smooths are constrained to have the same smoothing parameter, and if no variation over time is present, will be shrunk to random intercepts. The factor smooth for the interaction of time by speaker thus accounts for differences in the average pitch of individual speakers as well as for speaker-specific realization of intonation.  By including this factor smooth in the model, it is not necessary (nor desirable) to implement speaker normalization before statistical analysis.

For the influence of the tonal context on the target tone (which is denoted as \texttt{tone sequence} in the model specification), we included by-\texttt{tone sequence} random slopes.  We also considered an alternative model that includes  a factor smooth for the interaction of normalized time by tone sequence. However, the differences captured by this factor smooth were predominantly differences in the intercept. Furthermore, for our models of interest, prediction accuracy for held-out data was not better compared to a model with only random intercepts. Therefore, we decided to only include random intercepts for \texttt{tone sequence}.

The last line of the model specification requests random curves for \texttt{normalized time} in interaction with \texttt{X}. These smooths are random effect smooths \citep[see][for detailed discussion of such smooths]{baayen2022note}. 

In many of the following figures, we visualize the \textbf{partial} effect of one or more smooths. These partial effects show the effect of a predictor that is independent of all the other predictors in the model. In other words, these figures can be read as showing what a smooth term contributes to a pitch contour with all other predictors controlled for.

\section{Results}\label{sec:results}

\noindent
We first present the effects of the control variables. These effects are mostly independent of which predictor \texttt{X} is included, and are therefore reported for the baseline model. 
We subsequently report the central analyses focusing on the core predictors \texttt{tone pattern}, \texttt{word}, and \texttt{word sense}.

\begin{figure}[htbp]
    \centering
    \includegraphics[width=1\linewidth]{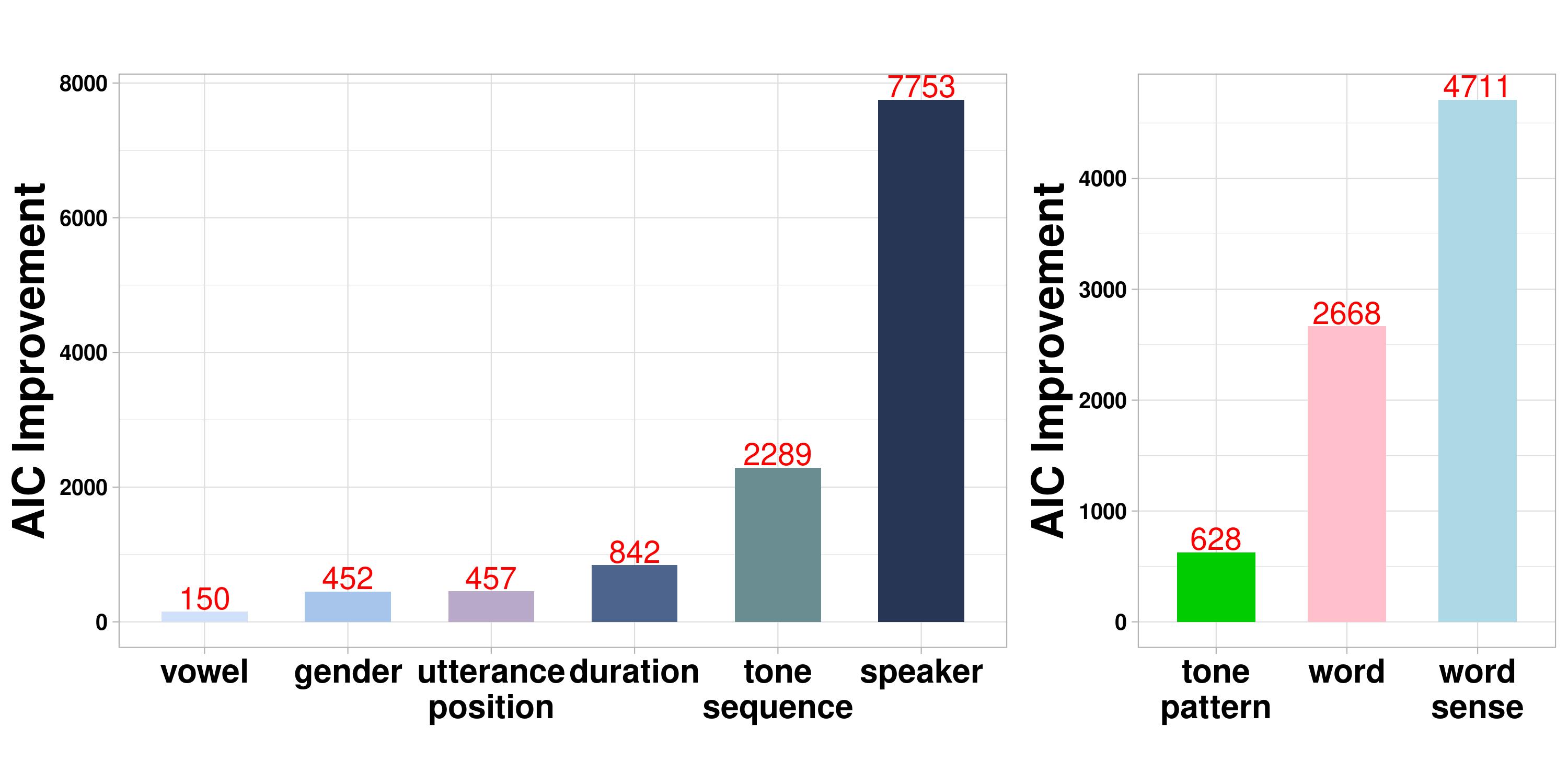}
    \caption{Importance of control variables (left panel) and core predictors (right panel). The left-hand panel shows decrease in model fit, gauged by increase in AIC units when a control variable is excluded from the baseline model. The right-hand panel shows the AIC units improvement when a core predictor is added to the baseline model. }
    \label{fig: AIC}
\end{figure}

All GAM models showed clear differences in pitch contours between speakers, and \textbf{speaker} has the highest variable importance in Figure~\ref{fig: AIC} ($\Delta$ AIC: 7753). Speaker also provides further fine-tuning of the difference in average pitch height within the two genders.   The next most important control variable is \texttt{tone sequence} ($\Delta$ AIC: 2289), which accounts for  tonal co-articulation.  The inclusion of \texttt{duration} and \texttt{gender} resulted in improvement of 842 and 457 AIC units, respectively. The main effect of gender indicates that,  on average, the pitch contours of female speakers are 0.39 F0 log units higher than the pitch contours of male speakers. 

\begin{figure}[H]
    \centering
    \includegraphics[scale=0.5]{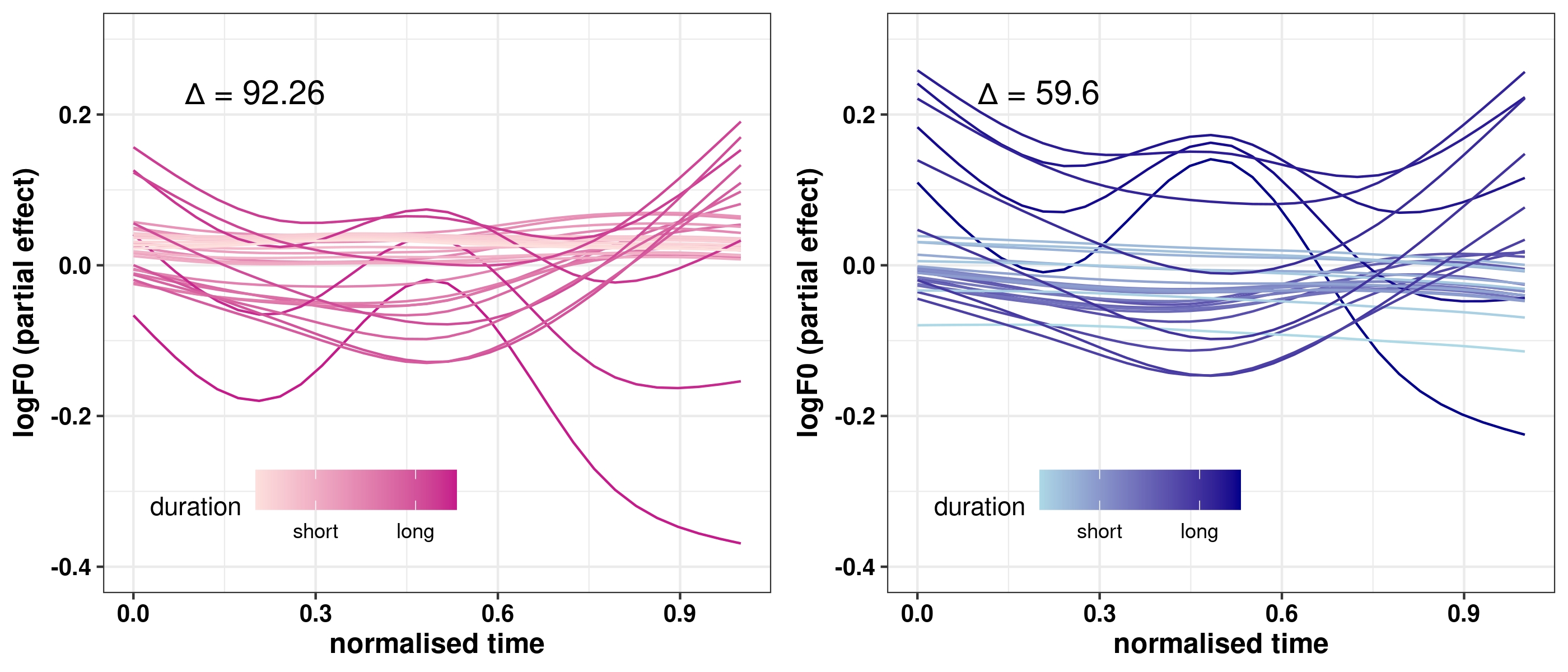}
    \caption{The partial effect of the interaction of log-transformed \texttt{duration} and  \texttt{normalized time} for female (left) and  male speakers (right). Darker colors indicate longer word durations. The number in the upper-left corner of each panel indicates the difference in Hz between the lowest and highest values, taking the intercepts for female and male speakers into account, and back-transforming from predicted log F0.}
    \label{figure:duration}
\end{figure}

The effect of \texttt{duration} on pitch was different for male and female speakers, as illustrated in Figure~\ref{figure:duration}, which is perhaps unsurprising given the sociophonetics literature on gender \citep[see, e.g.][]{podesva2014sociophonetics,baranowski2013sociophonetics,liang2011sociophonetic}.  In our dataset, female speakers contribute more tokens than male speakers (by-female average: 127, by-male average: 91), and the mean word duration for female speakers is slightly reduced compared to that of male speakers (mean females 0.092, mean males 0.096, $t_{(4364.3)} = -2.21, p = 0.02$).

Figure~\ref{figure:duration} shows that for shorter words  (lighter colors), the partial effect of \texttt{duration} on the pitch contours is almost flat and close to zero, for both genders, as expected given the literature on speech motor control \citep{perkell2000theory}.  For longer word tokens, the partial effect for females first falls,  then rises, and finally levels off.  For male speakers, the partial effect on the pitch contour resembles that of female speakers, but with a smaller range (range females: -0.36 to 0.19, range males: -0.23 to 0.26).

\begin{figure}
    \centering
    \includegraphics[scale=0.5]{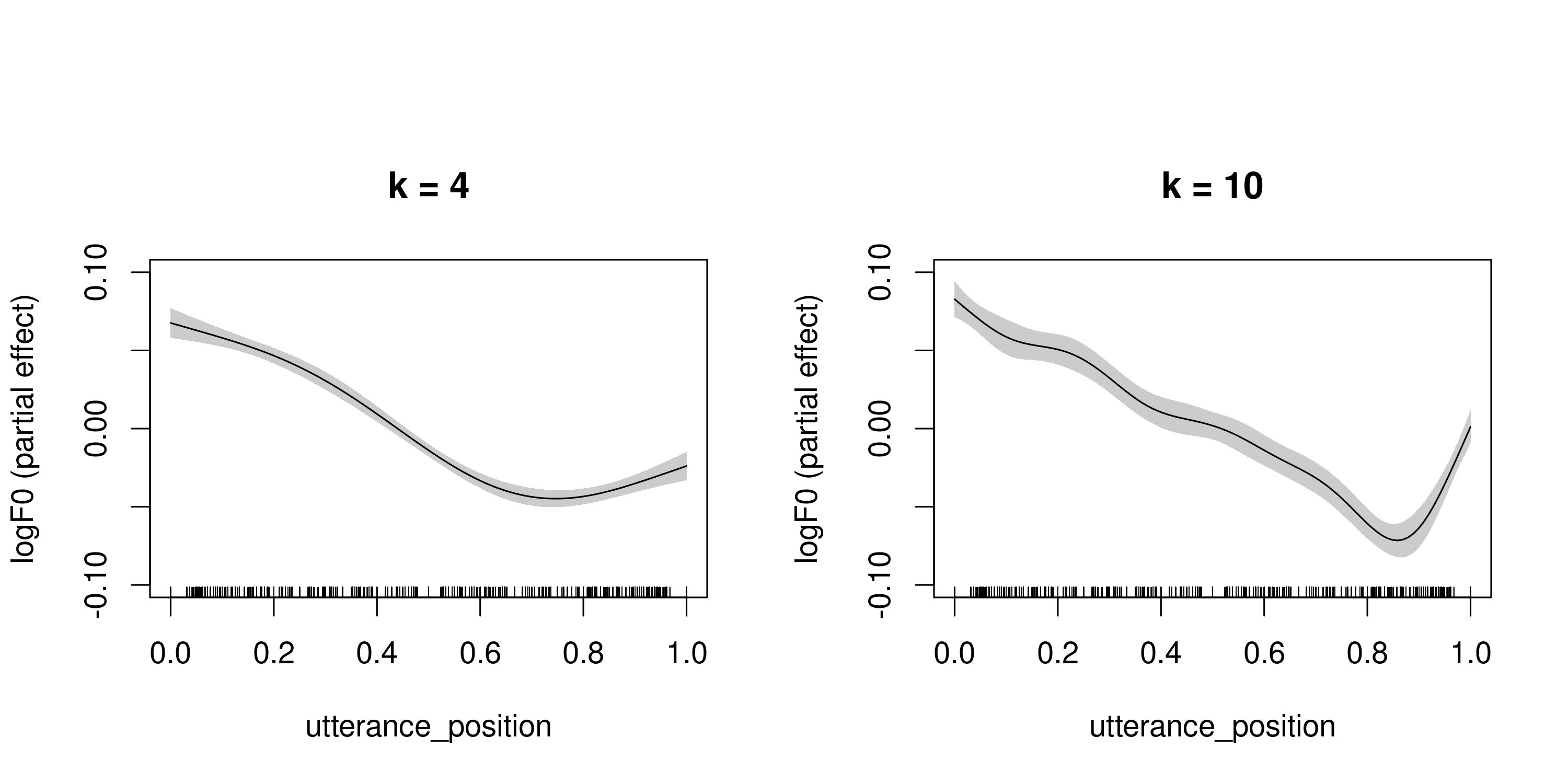}
    \caption{Partial effect of utterance position.  Words later in an utterance tend to be produced with lower pitch, irrespective of whether the number of basis functions is set to 4 (left panel) or to 10 (right panel).}
    \label{fig:parteffectPosition}
\end{figure}

Finally, vowel height and utterance position also contribute to the tonal variation, with 150 and 457 AIC units, respectively.   
The partial effect of utterance position is shown in Figure~\ref{fig:parteffectPosition}. As expected, words that occur later in an utterance are pronounced with lower pitch.  We modeled the effect of utterance position with a thin plate regression spline smooth with the number of basis functions $k$ set to 4, but this partial effect remains similar when $k$ is set to 10.  As the additional wiggliness for $k=10$ is not well interpretable, and as we aimed to keep the model relatively simple, the results that we report are based on the smooth with four basis functions.\footnote{
We did not implement an interaction of utterance position by normalized time, in order to avoid overfitting given that several words are positionally highly restricted.  For instance,  particles such as 吧, 啊, 嘛, and 啦 occur mainly in utterance-final position, whereas particles such as 的, 得, and 地} cannot occur utterance-initially.

\subsection{Tone pattern}
\begin{figure}[H]
    \centering
    \includegraphics[width=1\linewidth]{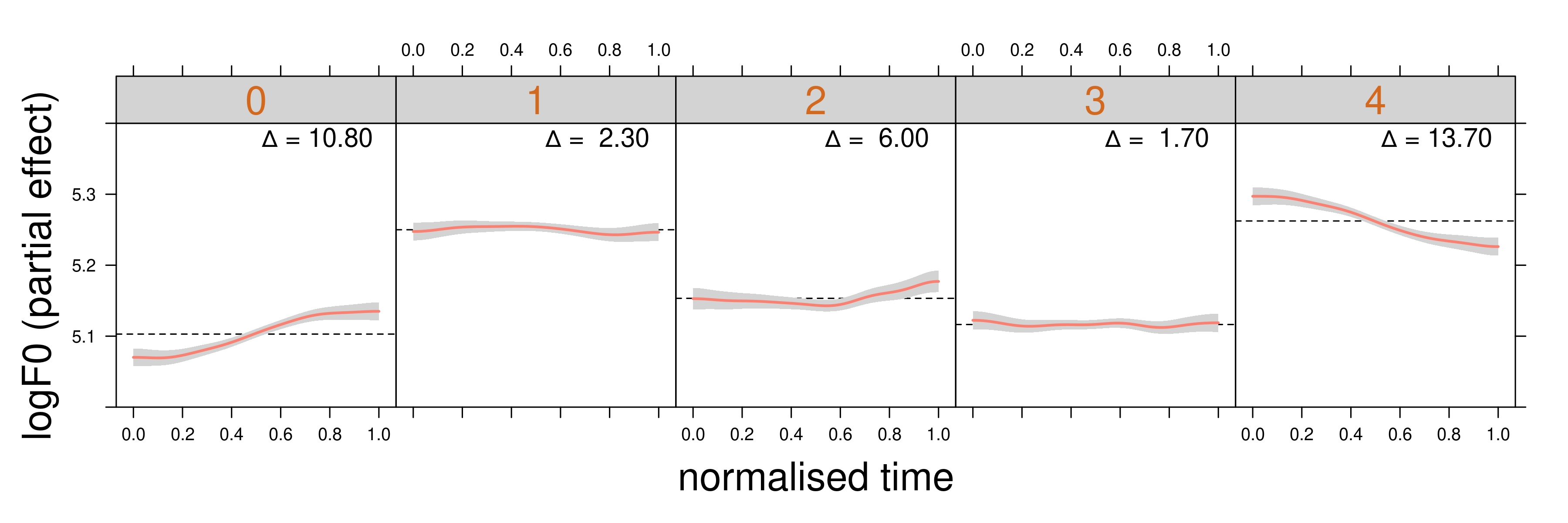}
    \caption{Partial effect of \texttt{tone pattern} in interaction with normalized \texttt{time}. Dashed lines denote the intercept for each tonal pattern predicted by the GAM. The number in the upper-right corner of each panel indicates the difference in Hz between the lowest and highest values.}
    \label{fig:tone pattern}
\end{figure}

Unsurprisingly, \texttt{tone pattern} improved model fit (see the green bars in the right panel of Figure~\ref{fig: AIC}, $\Delta$ AIC: 628). As shown in Figure~\ref{fig:tone pattern}, the partial effect of \texttt{tone pattern} is basically invariant over normalized \texttt{time} for T1, T2 and T3. Only T2 is not completely flat, and shows a small rise of 6 Hz, which is only just above the just noticeable difference \citep{jongman2017just}.
For T4, we observe the expected decrease in F0, as described in the literature \citep{hsu2009tonal}, and, for T0, an increase is visible,  which partially fits with the description of T0 given by \citet{huang2012study} as having a mid-low pitch target. For T4 and T0, the difference between the lowest and highest pitch values is about 14 Hz and 11 Hz respectively .

\subsection{Word-specific tonal contours}\label{sec: word-specific tonal contour}

\noindent 
Following up on the studies by \citet{Chuang:Bell:Tseng:Baayen:2024,lu2025realization}, we now consider a GAM that includes a non-linear random effect for the interaction of word by normalized \texttt{time}, using a factor smooth.  As our dataset comprises 95 words, this resulted in 95 word-specific partial effect pitch contours. The AIC scores ({pink bar}) shown in  Figure~\ref{fig: AIC} indicate that the model with a \texttt{word} factor smooth is a  substantial improvement ($\Delta$ AIC 2668) over the model with \texttt{tone pattern}, replicating for monosyllabic words the observation made by \citet{Chuang:Bell:Tseng:Baayen:2024} for disyllabic words with the T2-T4 tone pattern, and a similar observation by \citet{lu2025realization} for disyllabic words in general.

\begin{figure}[H]
    \centering
    \includegraphics[width=1\linewidth]{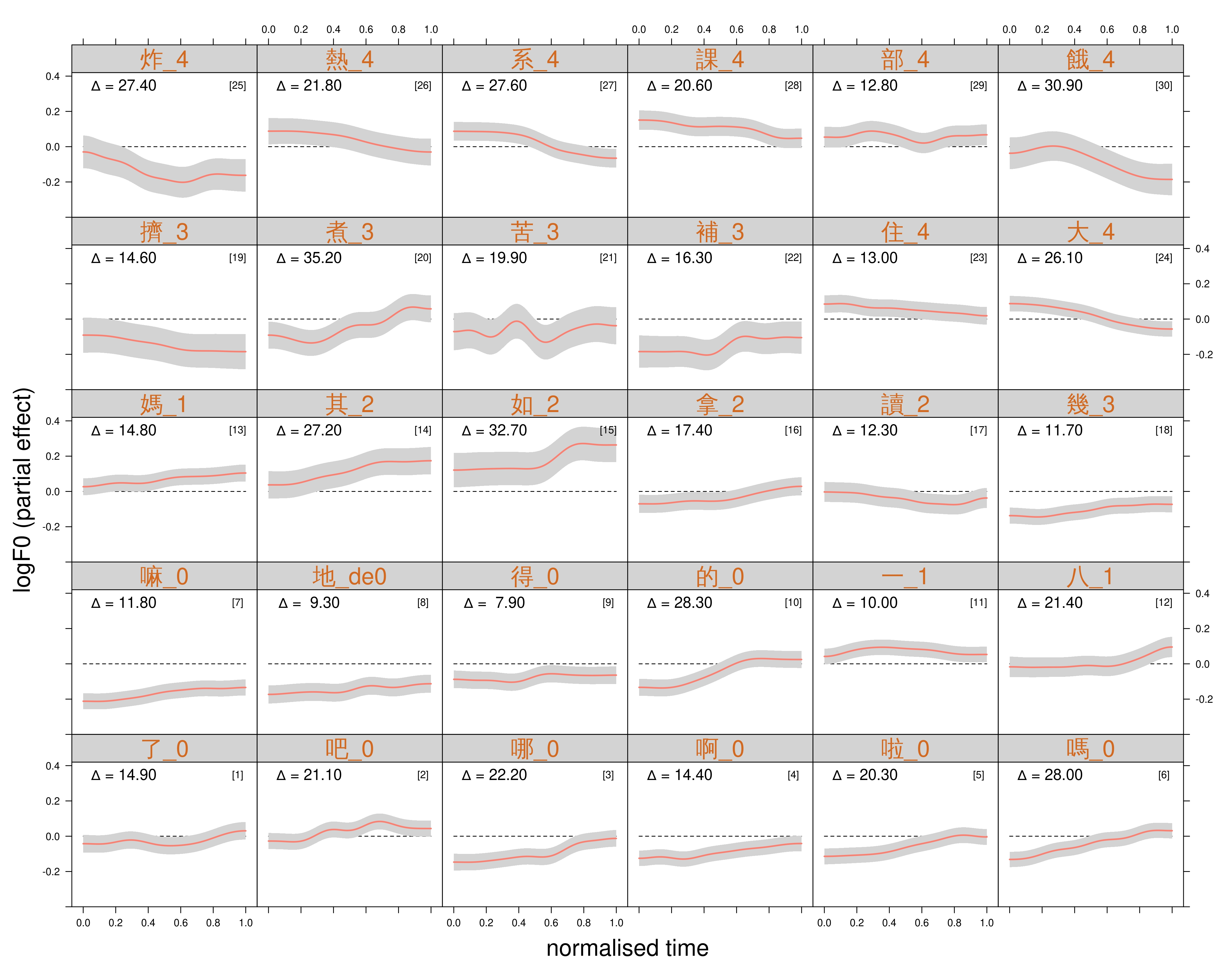}
    \caption{Word-specific pitch signatures for all words for which these signatures are well-supported statistically (15 function words and 15 content words).  Dashed lines denote the intercept for each tonal pattern predicted by the GAM. The number in the upper-right corner counts the panels. The number in the upper-left corner of each panel indicates the difference in Hz between the lowest and highest values when this partial effect is combined with the pitch intercept for female speakers.} 
    \label{fig:word}
\end{figure}

Figure~\ref{fig:word} presents those word-specific partial tone contours that are especially well supported.\footnote{
We made use of factor smooths to model random wiggly pitch contours for words, technically, an interaction of time by word.  This interaction is well supported, irrespectively of whether support is assessed on the basis of the model summary, or on the basis of the increase in AIC when the factor smooth for word is removed from the model specification.  However, knowing that there is an interaction of word by time is not informative about which words have large and robust pitch signatures, and which words have modest or negligible pitch signatures.  Just as in a linear mixed model, the posterior modes for a random intercept term will tend to be close to zero (as their distribution is assumed to be Gaussian), random wiggly curves will tend to mostly show minor deviations from the general pitch contour. But because we have multiple observations for each individual random wiggly curve, it is possible to evaluate more precisely whether a given curve is well-supported.  Therefore, to obtain insight into which words have strong signatures, we refitted the model with a by-smooth instead of a factor smooth \citep[for technical details on these two kinds of smooths, see][]{baayen2022note}. Importantly, the model summary for a by-smooth reports for each word whether its pitch signature is significant. We set $\alpha$ to 0.001 when determining the words for which a pitch signature is well-supported.}     

The partial effect size of these word smooths is within the (-0.2, 0.4) log F0 range. Estimated differences in Hz  range from about 8 to 36 Hz, which are all differences that are likely to be perceivable. These differences are greater than those induced by \texttt{tone pattern}.

\begin{CJK}{UTF8}{bsmi}
For most of the words with T0 (panel 1-10), a rising contour can be observed. The small rising pitch contour observed for the T0 (see Figure~\ref{fig:tone pattern}) is strengthened for the words with T0 in Figure~\ref{fig:word}. 嗎 \textit{ma0} (panel 6) and 的 \textit{de0} (panel 10) displays the strongest rising pattern with an increase of 28 Hz. In contrast, 嘛 \textit{ma0} (panel 7) and 地 \textit{de0} (panel 8) have somewhat lower pitch height compared to the other T0  words.

Panels 11-13 present the pitch contours for words with T1. Figure~\ref{fig:tone pattern} clarifies that T1 is basically a flat tone, similar in height to T2, but with a higher pitch than T3. The words 一 \textit{yi1} (panel 11), 八 \textit{ba1} (panel 12) and 媽  \textit{ma1} (panel 13) all have different contours. For 一, pitch rises and then falls; for 八, a late rise is present; for 媽, a steady rise is visible.

Panels 14-17 show words-specific modulations of T2, which is basically a flat mid tone in Figure~\ref{fig:tone pattern}. 其 \textit{qi2} (panel 14), 如 \textit{ru2} (panel 15) and 拿 \textit{na2} (panel 16) have rising patterns, which dovetails well with the canonical contour of T2, a rising tone. The word-specific pitch contour of 讀 \textit{du2} (panel 17) is basically flat, with a slight dip near the end of the vowel. 

Four words (panel 18-22) with the canonical dipping tone (T3) have their own tonal signatures, which depart from the low flat tone observed for the T3 tone pattern in Figure~\ref{fig:tone pattern}. Only 煮 \textit{zhu3} (panel 20) and 苦 \textit{ku3} (panel 21) have contours that could be interpreted as a dipping contour, but they appear to mainly have a rising tone. 擠 and  幾 \textit{ji3} have basically level tones, with a different intercept. For 補 \textit{bu3}, the pitch contour initially increases, and then levels off.\footnote{

We have checked for the 18 words in our dataset that carry T3 how often they are  followed by another T3 syllable to their right, the environment for T3-to-T2 tone sandhi. This turns out to be the case for 223 tokens.  These 223 tokens instantiate 5 tone sequences (e.g., 0-3-3 or 1-3-3).  For the same 18 words, we also extracted all tone sequences where the target word is not followed by T3 (e.g, 0-3-2 or 4-3-4). This second set of types was represented by 905 tokens.  Comparing the intercepts of the 905 tokens of tone sequences where the conditions for 3-3 sandhi are not met, and the 223 tone sequences where 3-3 sandhi is expected, no clear difference was found (means 5.16, 5.12, $t_{(8.2406)} = 0.87, p = 0.41$). In other words, there is no clear evidence in the present dataset for T3-T3 sandhi taking place for monosyllabic words with T3. As shown by \citet{lu2024form}, T3-to-T2 tone sandhi is present for two-syllable words in Taiwan Mandarin.
}

T4 has a falling pitch contour, as shown in Figure~\ref{fig:tone pattern}. Of the 8 T4 words with their own pitch signatures (panel 23-30), 7 words also have a falling pitch, indicating that these words have pitch contours with even more downward movement than is the case for T4 words in general.  In contrast, word 部 \textit{bu4} (panel 29) has flat contour.

Many of the words with the neutral tone are sentence-final particles (e.g., 嘛 \textit{ma0}) or can be sentence final (e.g., 了 \textit{le0}). For many of these words, Figure 5 shows a rising pitch contour, but the contours of these words can also be nearly flat (了 in panel 1). It is possible that the rising pitch contour is favored by particles in sentence-final position.  However, content words such as 媽 （\textit{ma1}), 拿 (\textit{na2}) and 煮 (\textit{zhu3}) (panels 13, 16 and 20) also have rising tone signatures. Unfortunately, our dataset is too sparse to allow for systematic exploration of interactions of normalized time by word by sentence position.  We leave the exploration of interactions of sentence position, more fine-grained indicators of prosody, and word category, to follow-up research. 

\end{CJK}

%
%
%
%

\subsubsection{Pitch contours for heterographic homophones}\label{sec: homo}

We have seen above that word co-determines the fine details of the pitch contours of monosyllabic words in Taiwan Mandarin, substantially outperforming \texttt{tone pattern} with respect to model fit. If it is indeed word meaning that is fine-tuning words' pitch contours, then heterographic homophones are expected to have their own pitch signatures.

\begin{sidewaysfigure}
 \centering
\includegraphics[width=1\textwidth]{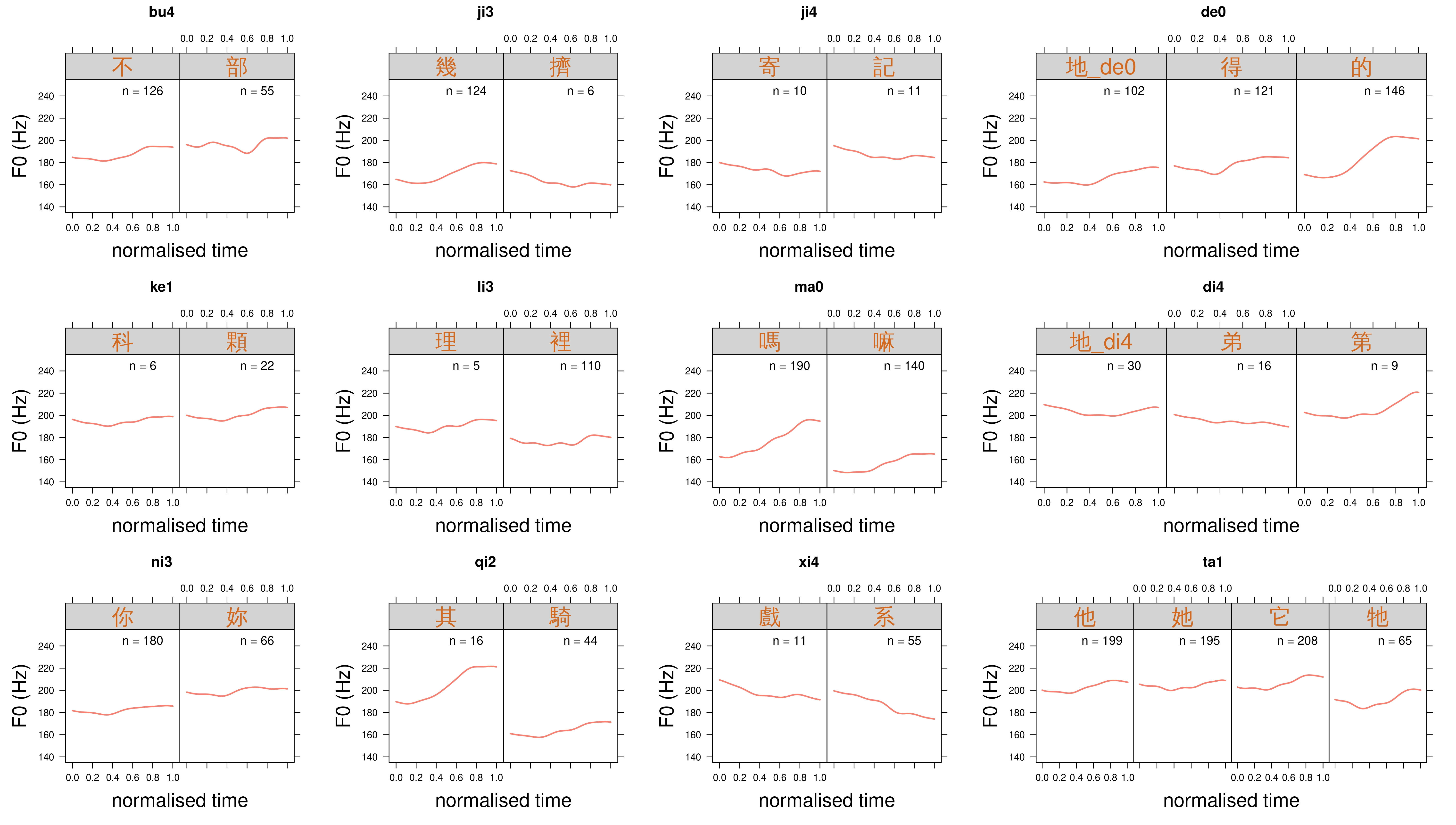}
\caption{The estimated pitch contours of heterographic homophones in our dataset. The horizontal axis represents \texttt{normalized time}, and the vertical axis predicted Hz. The numbers in the upper-right corners of the panels refer to the number of word tokens in our dataset.}
\label{fig:homo}
\end{sidewaysfigure}

\begin{CJK}{UTF8}{bsmi}
Our dataset contains 12 sets of heterographic homophones:  nine doublets (\textit{bu4}, \textit{ji3},\textit{ji4}, \textit{ke1}, \textit{li3},\textit{ma0}, \textit{ni3}, \textit{qi2},       \textit{xi4}), two triplets (\textit{de0}, \textit{di4} ), and a quadruplet (\textit{ta1}). The pitch contours in Hz for a female speaker predicted by the \texttt{word} GAM for these homophones  are visualized in Figure~\ref{fig:homo}. 
For prediction with the GAM, \texttt{Speaker} was set to the first female speaker in the dataset. \texttt{Duration} was set to the medium value in the dataset. \texttt{Utterance position} was set to 1,  with the adjacent tone pattern fixed at NULLNULL (preceding and following pause) for T1,T3 and T4 and for T0 and T2 was set to T0 T0 (preceding and following neutral tone). 

There are clear differences between heterographic homophones visible.
The negation  不 has a pitch contour that is relatively flat, whereas its heterographic homophone 部 (`section'/`department') has a dipping tone pattern. The verb 擠 \textit{ji3} `squeeze' and the function word 幾 \textit{ji3}, `several' also have somewhat different pitch contours: 幾 has a dipping curve, compared to the flat contour observed in 擠. The words  寄 \textit{ji4} `send' and 記 \textit{ji4} `remember' differ mainly in pitch height. Neither word has a convincing dipping tone. 

The first right panel of Figure~\ref{fig:homo} presents  the pitch contours for three homophonic function words, all pronounced as \textit{de0}.  The character 地,  when realized with T4 and with /i/, denotes `ground'. (This character also appears again in the middle right panel for \textit{di4}, but there it is pronounced as \textit{de0}.)  This function word appears in constructions where it is preceded by a verb or adjective, and followed by another verb, and realizes a meaning similar to English adverbial \textit{-ly}. Its homophone  的 is used both as genitive marker and as complementizer. The third member of this homophone triplet,  得, is also an adverbial marker that appears in constructions where it is preceded by a verb or adjective, and followed by an adverb or adjective.  Even for native speakers, selecting which of the three characters to use in writing is far from trivial, and in mainland China, spelling rules for less formal registers have recently been relaxed and allow the genitive marker 的 to be used across all constructions.  Nevertheless, there appear to be some differences in how these three words are pronounced.  The contour for 的, showing a strong upward trend, contrasts with the contours of the other two function words, whose contours are more level. 

 With respect to the pitch contours for the homophones of \textit{ke1}, 顆 `measure word' has a perhaps slightly higher pitch contour than 科 `branch'/`subject'. 裡 \textit{li3} `inside' has a lower overall pitch compared to 理 \textit{li3} `organize'. For 嗎 \textit{ma0}, question marker and 嘛 \textit{ma0}, sentence final particle,  a stronger rising pitch is visible in 嗎. The middle right panel presents the pitch contours for 地 \textit{di4} `ground',  弟 \textit{di4} `brother' and 第 \textit{di4} `representing order'. The kinship term is realized with a falling tone while 第 is realized with a rising tone.
 
The phonological word \textit{ni3} is represented in Taiwan Mandarin by two characters:  妳 for female second person singular and  你 for male second person singular. The pitch of 妳 has a higher onset than  你. The verb 騎 \textit{qi2} `ride' has a lower pitch height and a flatter, hardly rising, pattern than its homophone, the function word 其 \textit{qi2} `he/his/she/her, referring to somebody mentioned before'. For the last doublet \textit{xi4}, the expected falling tonal pattern is observed in 系 `system'/`group' and 戲 `drama'/`show', while the tonal realization for 戲 is relatively higher in height.

The lower right panel of Figure~\ref{fig:homo} concerns four heterographic homophonic personal pronouns, all of which share the same canonical high tone and denote third  person singulars:   她 `she',  他 `he',  它 `inanimate it', and 牠 `animate it'. The latter character is specific to Taiwan Mandarin, as in standard Mandarin, only  她,  他, and  它 are in use. Compared to more frequent 她,  他, and  它, 牠 has a lower overall pitch height and has a clear rise that starts in the first half of the vowel.
\end{CJK}

\subsection{Word sense-specific tonal contours}\label{sec: word sense}

The results of the preceding section support the possibility that word meaning is co-determining the way in which, in Taiwan Mandarin, pitch contours are realized. In this section, we consider whether we can replicate the results of \citet{Chuang:Bell:Tseng:Baayen:2024} and \citet{lu2024form}, who reported that model fit improved when \texttt{word} factor smooths for \texttt{normalized time} are replaced with \texttt{word sense} factor smooths. Their findings, as well as the present results for heterographic homophones, strongly suggest that meaning is an important co-determinant of words' F0 contours.  
\begin{CJK}{UTF8}{bsmi}
We thus examined a GAM in which \texttt{word sense} replaced \texttt{word}. The word senses of individual word tokens were estimated using the method described in \citet{Chuang:Bell:Tseng:Baayen:2024}, which makes use of the Chinese Wordnet and BERT-derived contextualized embeddings
\citep{huang2010constructing,hsieh2020tutorial}.  By way of example, this method assigns the word sense `to read' to  讀 \textit{du2} in  讀漢書 \textit{du2 han4-shu1} `read History of the Former Han' or  in 讀史記 \textit{du2 shi2-ji4} `read Records of the Grand Historian', but the word sense `to study' when  讀 \textit{du2} is used in the phrases  讀六年級 \textit{du2-liu4nian2ji2} `study in sixth grade' or in  讀小學 \textit{du2-xiao3xue2} `study in elementary school'. We note here that the theoretical concept of `word sense', although intuitively attractive, is difficult to apply in practice. As pointed out by \citet{kilgarriff2006word}, there are `no decisive ways of identifying where one sense of a word ends and the next begins'.  Polysemy is actually much more subtle and nuanced than a set of discrete sense classes would suggest. Therefore, since sense labeling can only be approximate, models working with sense rather than word are tentative. Below, we therefore complement the  analyses using word senses with analyses that make use of contextualized embeddings to approximate word meaning in context.
\end{CJK}

Figure~\ref{fig: AIC} (presented above) presents the improvement in AIC when word sense (blue bars) is added to the baseline model.  We observe a substantial improvement in model fit compared to the baseline model (by 4711 AIC units).  Overall, \texttt{word sense} emerges as the best predictor, outperforming both \texttt{tone pattern}, and \texttt{word} in GAM models.

\begin{figure}
    \centering
    \includegraphics[width=1\linewidth]{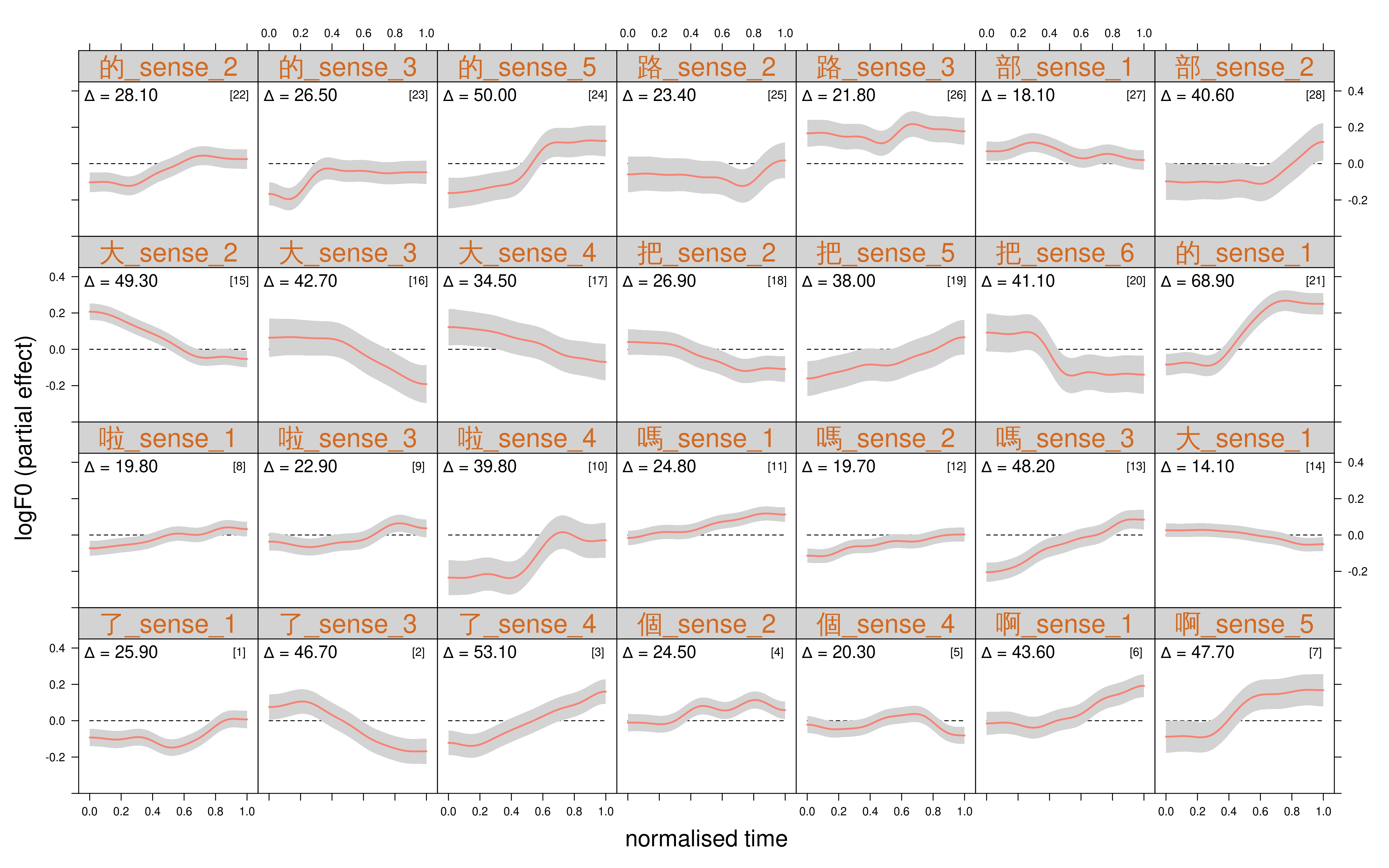}
    \caption{Selected partial effects of \texttt{word sense}. The number in the upper-left corner of each panel indicates the difference in Hz between the highest and lowest values in the pitch contours. These differences were calculated for  female speakers. For English translations, see Appendix~\ref{appendix translation}.}
    \label{fig:sense}
\end{figure}

Figure~\ref{fig:sense} displays the partial effect of \texttt{word sense} for the subset of words for which more than one clear word sense-specific contour is detected.  
\begin{CJK}{UTF8}{bsmi}
Many of the function words in our data set, such as the sentence final particles  了 \textit{le0}, 啊 \textit{a0}, 啦 \textit{la0}, and 嗎 \textit{ma0} have multiple word senses. These function words are all associated with T0 as canonical tone. However, a rising tonal pattern is observed for 了 (panels 1 and 3), 啊 (panels 6 and 7), 啦 (panel 8 and 10), and 嗎 (panels 11-13).

個 \textit{ge0} (panels 4-5) is a measure word. The noun following it can be a person (panel 4). The referent can also be unspecified, as in 那個 \textit{na4-ge0} `that one' (panel 5). These different uses of 個 go hand in hand with differences in pitch realization.
\end{CJK}

大 'big' \textit{da4}, can carry the meaning of `something is greater than the object being compared', such as 很大支持 \textit{hen3-da4-zhi1-chi2} 'big support' (panel 14). In this meaning, it has relatively flat tone. 大  can also realize the meaning `bigger in capacity' such as 大禮物 \textit{da4-li2-wu4} `big present'. As illustrated in panel 15, in this case, its pitch contour is realized with the expected falling tone, with a final levelling of F0. 大 in the sense of `elder/older' (panel 16), 我爸爸大很多歲 \textit{wo3-ba4-ba4-hen3-duo1-sui4} `My dad is much older', is realized with a late expected falling tone. In panel 17, 大 with the meaning of 'old' 年紀太大 \textit{nian2-ji4-tai4-da4} 'too old' has a falling contour. As these word senses are less-attested in our dataset, the confidence intervals of pitch contours are wide, and pitch lowering is less visible than for the word sense `something is greater than the object being compared' (panel 14). 

\begin{CJK}{UTF8}{bsmi}

把 (\textit{ba3}) can be used as a measure word or a verb. In panel 18, 把 represents the  measure word, and is always preceded by a numeral, as in 一把工作 \textit{yi1-ba3-gong1zuo4}, `a bunch of work'. Its contour is basically flat with a final lowering.  When 把 is used as a verb  (panel 19), it can mean using the hand to take or hold things (e.g.,  把個羽球 \textit{ba2-ge0-yu3qiu2} `hold the badminton shuttle'). In 把這些書帶回去 (panel 20),  \textit{ba3-zhe4-xie1-shu1-dai4-hui2-qu4} (`take these books back'), 把 is the marker for the direct object.  With the former word sense (panel 19),  把 is realized overall with a rising contour, while with the latter word sense (panel 20), a falling contour is observed that, however, levels off near the end of the vowel.

\end{CJK}

To summarize, considered jointly, the results of the GAM models support the importance of \texttt{word sense} as co-determinant of tone contours, showing that the findings of \citet{Chuang:Bell:Tseng:Baayen:2024} and \citet{lu2024form} are not restricted to disyllabic words. 

\subsection{Cross-validation}

In order to ascertain whether the present results are robust, we carried out a cross-validation study. We held out 10\% of the current data as test data, and used the remaining 90\% as training data. Every \texttt{word type} was represented in both the training and test data. The number of tokens per word in the test data was proportional to that in the training data. We fitted three models to the training data for 30 random held-out datasets: the baseline model, the \texttt{word} model and the \texttt{word sense} model. To quantify model accuracy, we obtained the models' predictions for the F0 contours of the held-out test data, and calculated the sum of squared errors (SSE). 

The \texttt{word} GAM model offered increased prediction accuracy over the baseline model (p=0.0001), replicating the findings of \citet{Chuang:Bell:Tseng:Baayen:2024} for disyllabic words with the T2-T4 tone pattern. The prediction accuracy of the \texttt{word sense} GAM model was indistinguishable from that of the baseline model, which also replicates the study by \citet{Chuang:Bell:Tseng:Baayen:2024}. Furthermore, the accuracy of the \texttt{word sense} model is worse than that of the \texttt{word} model (p<0.0001). 

There are several possible reasons for why the performance of the \texttt{word sense} model is worse than that of \texttt{word} model. First, monosyllabic words are highly polysemous, which challenges sense tagging and the reliability of \texttt{word sense} as predictor in the GAM. Second, from a statistical perspective, reduced accuracy of the \texttt{word sense} model for held out data is unsurprising:  the word senses are supported by fewer word tokens than the words (characters), which is especially problematic for word senses with relatively few tokens, training on even fewer tokens \citep[see also][for detailed discussion]{Chuang:Bell:Tseng:Baayen:2024}.

\subsection{Predicting pitch contours from contextualized embeddings}\label{sec: DLM}

Because the cross-validation results for word sense are inconclusive,
in order to provide stronger support for the hypothesis that words' semantics co-determine how their tones are realized, we tested whether the Discriminative Lexicon Model (DLM) \citep{baayen2019discriminative,chuang2021discriminative,Heitmeier:Chuang:Baayen:2025} can predict words' pitch contours on the basis of their meanings, which are quantified with a semantic distributional model.  Across a range of languages, the DLM has been used successfully to capture the alignment between meaning and fine-grained phonetic variation, such as the degree of tongue lowering in the articulation of the vowel /a/ in German \citep{saito2022articulatory}, the spoken word durations of homophones in English \citep{gahl_time_2024}, and tonal realization  in Mandarin disyllabic words \citep{Chuang:Bell:Tseng:Baayen:2024,lu2025realization}.  The DLM represents words' forms and meanings with high-dimensional numeric vectors, and defines mappings that predict meaning vectors from form vectors (comprehension), and form vectors from meaning vectors (production).  According to the DLM, it should be possible to predict words' pitch contours from their meanings. If this prediction is correct, this provides strong support for the effect of word being a semantic effect. 

In what follows, we use context-sensitive embeddings from distributional semantics to represent words meanings. The contextualized embeddings that we used were obtained with GPT-2 (for technical details, see \citet{Chuang:Bell:Tseng:Baayen:2024} and CKIP \footnote{ckiplab/gpt2-base-chinese, which is available on https://github.com/ckiplab/ckip-transformers.}). GPT-2 was applied to the utterances in the Taiwan Mandarin spontaneous speech corpus \citep{fon2004preliminary}, resulting in token-specific semantic vectors known in computational linguistics as contextualized embeddings.

Following \citep{Chuang:Bell:Tseng:Baayen:2024,lu2025realization}, we make use of a linear mapping from the matrix of contextualized embeddings (with one unique embedding for each token) to the matrix with the corresponding pitch contours.  
For a linear mapping from a matrix with contextualized embeddings to a matrix of pitch vectors, the pitch vectors need to have a fixed length. We obtained 100-dimensional pitch vectors from GAM models by having these models predict pitch at 100 equally-spaced points in normalized time.  At this point, the question arises what the optimal GAM is for obtaining 100-dimensional predicted pitch vectors. 

Pitch contours inevitably contain measurement noise that likely has multiple origins: noise in the speaker's production, noise in the audio recordings, noise in the word boundaries provided by the corpus, and noise due to our pitch extraction method.  When a predictor is relevant but withheld from a GAM, its effect will be merged with the by-observation noise. The resulting GAM will therefore provide predictions that are less precise. In what follows, we considered three GAMs that implement models with increasing amounts of prediction accuracy.

The first GAM model that we used to denoise the data and obtain 100-dimensional predicted pitch vectors was the baseline model.  This baseline GAM provides the lowest amount of prediction accuracy (model 1, AIC = -276052.4, R-squared 0.491). 

The second GAM model enriched the baseline model with \texttt{tone pattern} as additional predictor. This second GAM provides improved improved accuracy  (AIC = -276679.9,  decrease in AIC: 628, R-squared 0.492). The corresponding  100-dimensional predicted pitch vectors should be more precise.

The third GAM added \texttt{word} as additional predictor to the baseline model.  As this model provides the best fit to the data (AIC = -278720.5, difference in AIC with respect to model 1: 2668; R-squared=0.510), the 100-dimensional pitch vectors obtained with this model should have the highest quality. We expect that contextualized embeddings predict  pitch contours most precisely in the case of pitch contours obtained with the GAM that has access to word. If this expectation is borne out, then this provides further evidence for the importance of word for understanding Mandarin tonal contours.

We evaluated the quality of the linear mappings from contextualized embeddings to denoised pitch contours using a permutation baseline.  
For each of 30 runs, the data was split into 90\% training data and 10\% test data, in such a way that the number of tokens of any word type was proportionally represented in both the training and the test data.
A predicted pitch contour was evaluated as being predicted accurately if and only if its closest pitch contour neighbor (evaluated with the sum squared error) was associated with a token of the same word type. 

This procedure yields 30 accuracy values for the observed, empirical data. To assess whether these accuracies are above what can be expected under chance conditions, we considered two baselines.  The first baseline is a simple majority baseline obtained by always selecting the most frequent word type as prediction. This majority baseline is 3.4\%, the relative frequency of the most frequent word type in the data.

A permutation baseline was obtained by randomly permuting the word identifiers, and repeating 30 times the procedure of training on 90\% of the data and evaluating on 10\% of the data. The resulting permutation baseline (averaged across 30 runs) was 2.1\% (range 0.9\% to 3.4\%).

\begin{figure}[htbp]
    \centering
    \includegraphics[width=1\linewidth]{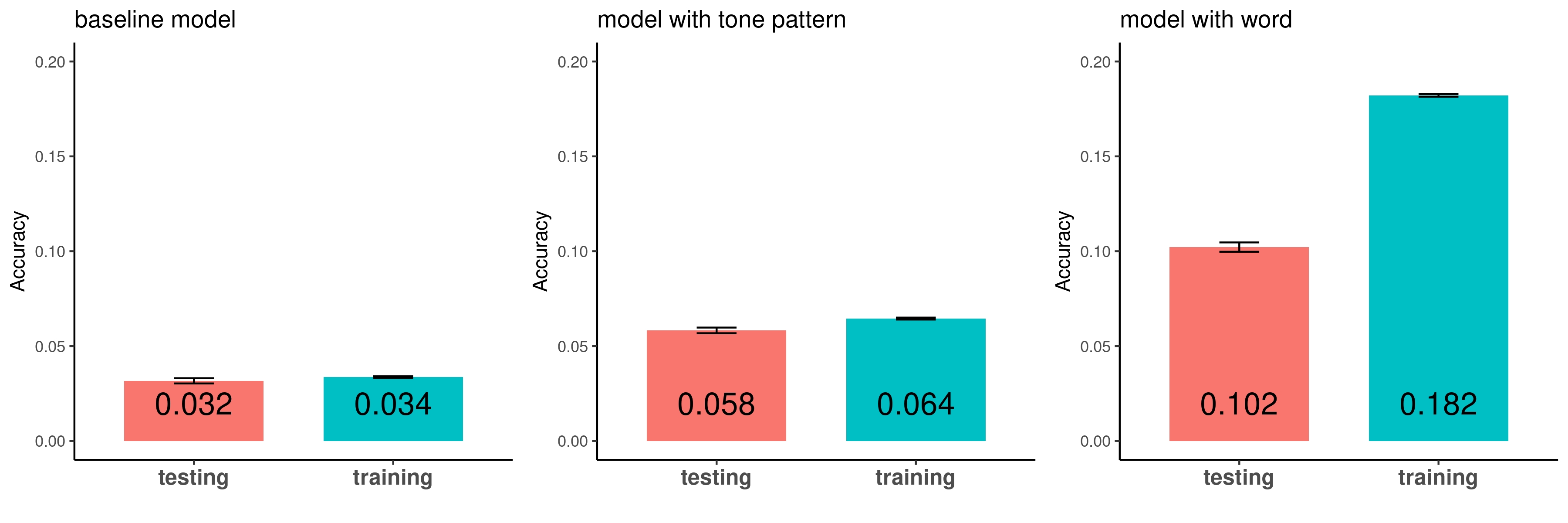}
    \caption{Mean production accuracies of DLMs  predicting the pitch contours (vectors) in training data and testing data which were denoised with  three different GAM models. From left to right: minimal denoising using the  baseline model, improved denoising using a GAM that adds \textbf{tone pattern} as predictor to the baseline model, and optimal denoising using a GAM that enriches the baseline model with \textbf{word} as predictor. Mean accuracy is obtained from 30 random training and testing splits, each trained and evaluated independently. Error bars indicate double the standard error.}
    \label{fig: LDL}
\end{figure}

Figure~\ref{fig: LDL} clarifies that the accuracy of the DLM model with pitch contours obtained from the baseline GAM, for both testing and training, is close to the majority baseline, and only slightly higher than the permutation baseline.  Pitch contours obtained with a GAM that also contains \texttt{tone pattern} as predictor are predicted by the DLM with a higher accuracy:   6.4\% for training and 5.8\% for testing.  This is well above both baseline accuracies.  Accuracy improves substantially for pitch contours denoised with the baseline GAM model enriched with word: 10.2\% for testing and 18.2\% for training.  As anticipated, the pitch contours from which most noise has been removed can be predicted most accurately from their contextualized embeddings.  These results replicate similar results reported by \citet{Chuang:Bell:Tseng:Baayen:2024} for bisyllabic words with the rise-fall tone pattern, and by \citet{lu2025realization} for all 20 tone patters for a restricted set of tonal contexts.

In summary, we have shown that the pitch contours realized on the word tokens in our dataset can be predicted from their corresponding contextualized embeddings, with an accuracy that exceeds a randomization baseline.
This result supports our hypothesis that the word-specific pitch contours that we have observed arise as a consequence of differences in words' contextualized meanings: meaning and pitch realization are entangled at the level of individual tokens.

\section{General discussion} \label{sec:gendisc}

\noindent
This study investigated the realization of the F0 contours of monosyllabic words in spontaneous conversations in Taiwan Mandarin.  A central finding is that the tonal realization of monosyllabic words is in part word- and meaning-specific.  

First, the word senses of monosyllabic Mandarin word tokens (determined on the basis of the context in which they occur, with a large language model) enable more precise prediction of their pitch contours than their word types (section~\ref{sec: word sense}). For example, as different pitch signatures when used with the sense of taking/grasping from when used as measure word for counting hand movements (see Figure~\ref{fig:sense}).

\begin{CJK}{UTF8}{bsmi}
Second, we show that the heterographic homophones in our dataset tend to have distinct F0 contours (section \ref{sec: homo}). For example, 你 (you, masculine) and 妳 (you, feminine) have different pitch signatures. 
\end{CJK}
Third, we also showed (using a cognitively motivated computational model to which we return below) that the pitch contours of individual word tokens can be predicted well above a permutation baseline from the corresponding contextualized embeddings (section \ref{sec: DLM}), which are approximations of word tokens' meanings in context. 

\begin{figure}[htbp]
    \centering
    \includegraphics[width=1\linewidth]{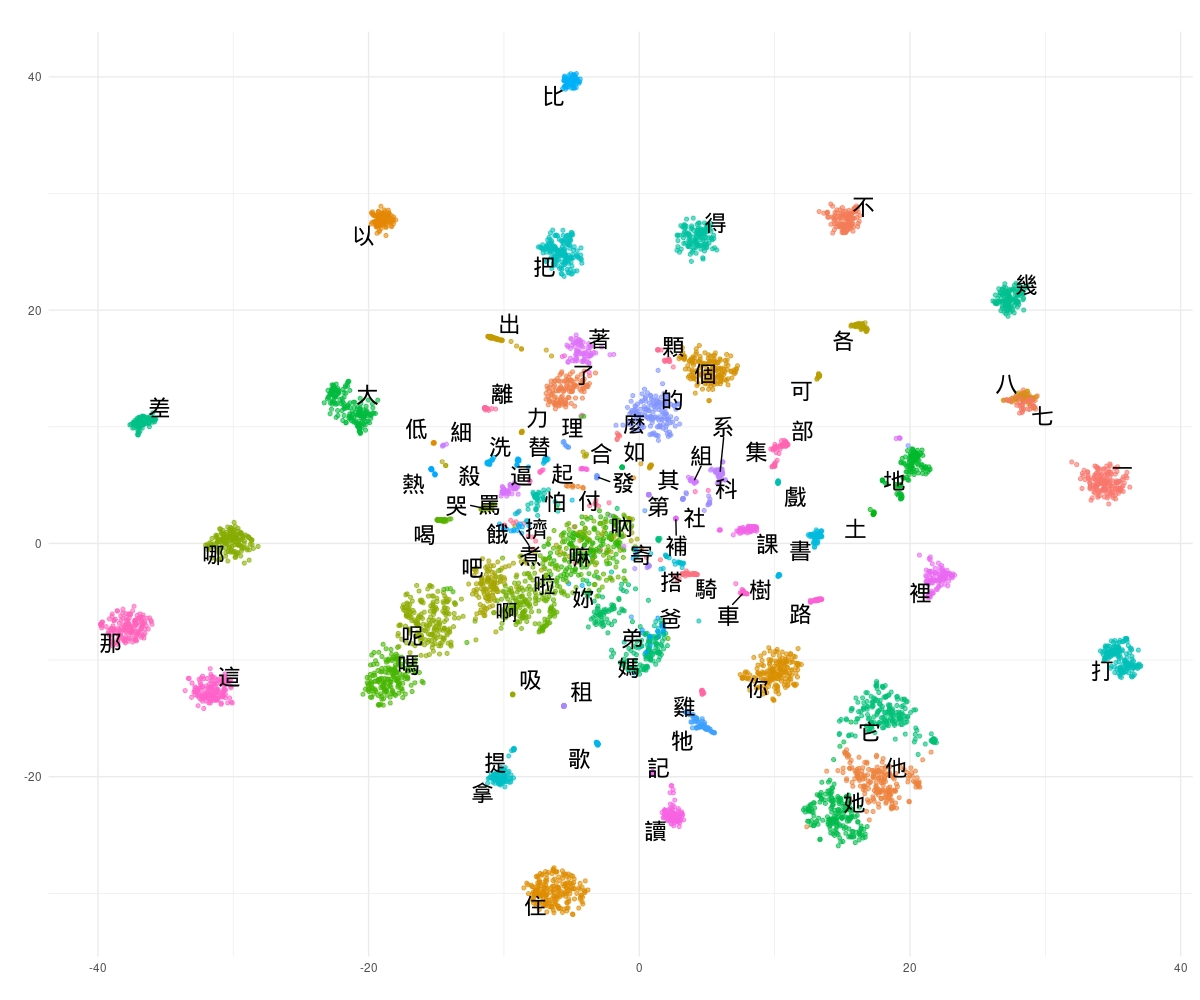}
    \caption{Contextualized embeddings, obtained from a pre-trained Chinese GPT-2 model, are shown in a two-dimensional plane obtained with t-SNE.}
    \label{fig:tSNE}
\end{figure}

It is known that  contextualized embeddings also  capture substantial amounts of syntactic information \citep{perez2021evolution}.  This may help explaining why contextualized embeddings are predictive for the phonetic realization of Mandarin pitch contours.  However, as can be seen in Figure~\ref{fig:tSNE}, obtained with an unsupervised clustering method \citep[t-distributed neighbor embedding][]{van2008visualizing}, the contextualized embeddings of our word tokens cluster primarily by word, i.e., they primarily capture words' meanings.  In the light of these three observations, it is clear that word meaning is a predictor that, although new to phonetics, has considerable explanatory value.
\begin{CJK}{UTF8}{bsmi}
In Figure~\ref{fig:tSNE}, semantically similar words clusters together in the tSNE map. The pronouns 她 `she', 他 `he', and 它`inanimate it' are together in the lower right, the verbs 拿 `take' and 提 `put' are adjacent, the numerals 八, 七 and 一 cluster in the upper right, the adjectives 大 'big', `tall' and 低 `low' are close together in the upper left, the associates 熱 `heat' and 喝 `drink' are also found in each other's vicinity, the demonstratives 那 `that' and 這 `this' are together in the lower left. Family members (媽 `mother', 爸 `father', 弟 `brother') are found to the left of the third and second person pronouns, with some other concrete nouns to their upper right (車 `car', 路 `road', 樹 `tree', 書 `book', and 課 `class'). The question particles 嗎 and 呢 are also placed near each other by the t-SNE algorithm. Function words and particles are spread out across the whole tSNE map, nouns tend to occur to the lower left of the origin, verbs are more spread out (住 `live' at center bottom,  把 `take' at center top,  the light verb 打 `hit' at the center right). 
\end{CJK}

These findings raise the question of why word- and meaning-specific tonal signatures exist, and how they arise.  To answer these questions, we considered several possible explanations.

One explanation is that the word effect is a prosodic effect. We therefore included words' positions in their utterance as a control variable. Although a word's position does not do justice to the full range of prosodic effects, it does capture an important aspect of prosody. Accordingly, the position of a word in an utterance has some explanatory value --- in general, pitch declines with time --- but its effect is relatively independent of the effect of word.
%
%
Word duration has also been argued to be prosodic in nature.  As expected, word duration is predictive for the realization of pitch, but the effect of word duration is also small compared to the effect of word.  In the light of these findings, we conclude that the effect of word is unlikely to be completely confounded with prosody. 

\begin{figure}
    \centering
    \includegraphics[width=1\linewidth]{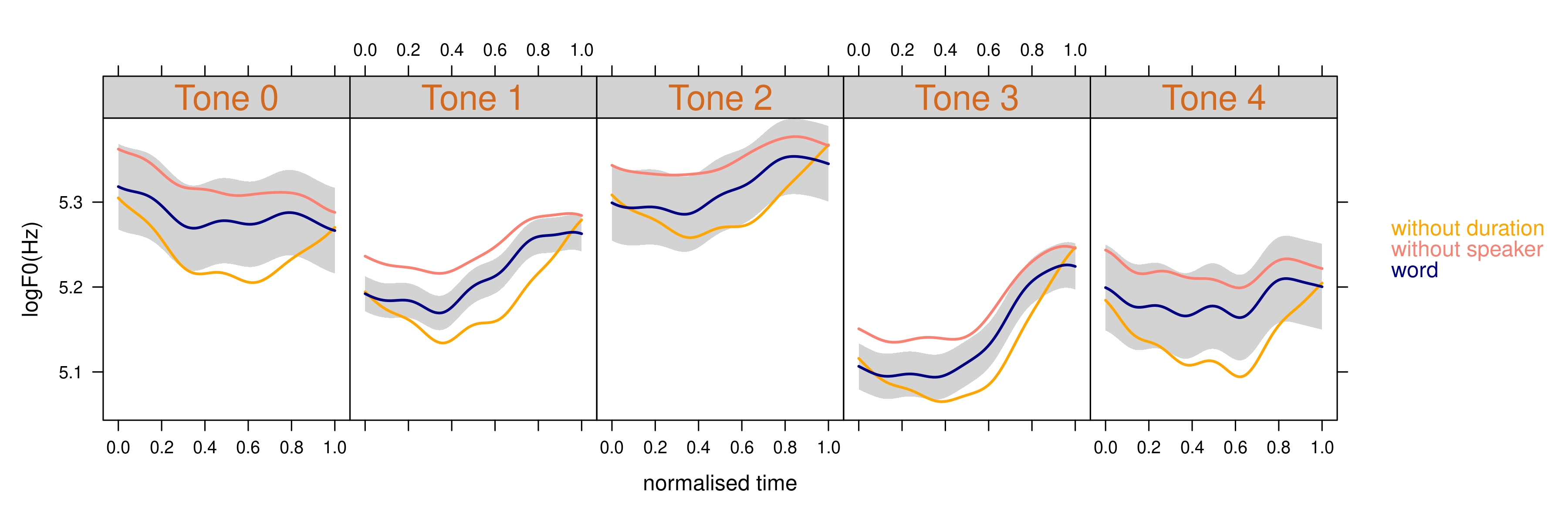}
    \caption{Predicted contours from a word-specific GAM model (blue line) with its confidence interval, and predicted contours from a GAM without \texttt{speaker} (pink line) and a GAM without \texttt{duration} (orange line). Word effects are shown for 呢 (T0), 它 (T1), 拿 (T2), 裡 (T3) and 住 (T4).  
    }
    \label{fig:word_duration_speaker}
\end{figure}

The robustness of the word effect is also supported by the fact that the shape of the word effect is relatively robust across different GAM models.  Figure~\ref{fig:word_duration_speaker} presents the partial effect of a GAM with word, duration, and speaker (blue), a GAM with word but not speaker (red), and a GAM with word but not duration (orange).
Importantly, across all model variants, the shape of the word effect is similar. This leads to the conclusion that there is a strong main effect of word that is fine-tuned but not fundamentally changed by other covariates.

Yet another explanation is that the word effect is actually an effect  of tonal co-articulation or of tone sandhi.  In our statistical models, we included a predictor for of the three tones consisting of the preceding tone, the current tone, and the following tone. 
This predictor has substantial variable importance. Importantly, also when tone sequence is included as a predictor, the effect of \texttt{word} is present, and, moreover, is larger than the effect of tone sequence. In other words, the effect of \texttt{word} cannot be explained as just being the result of tonal co-articulation.


Another explanation, especially for function words in our dataset, holds that the effect of word reflects differences in the syntactic constructions in which the words are used,  which have been reported to also co-determine the realization of tone \citep{duanmu2005tone,wang2020interaction}. As our data are extracted from a corpus of conversational speech, a register that is very different from the register of formal written language, assessing whether the effect of word is confounded with syntactic structure is far from straightforward, especially as there is no treebank for our corpus  that would allow for reliable automatic syntactic annotation.  In the studies by \citet{gahl2008time} and \citet{gahl2024time} on the spoken word duration of homophones, the probability of a word given the preceding word, and the probability of a word given the following word, were included as predictors intended to control for syntactic probabilities. A similar approach was used in the studies by \citet{Chuang:Bell:Tseng:Baayen:2024} and \citet{lu2025realization}. 
 We added these forward and backward probabilities as additional predictors, and implemented tensor product interactions with normalized time.  Model fit improved, but in cross-validation with 10 runs, prediction accuracy decreased significantly ($t_{(9)}=-5.161, p = 0.0006$).  Importantly, the effect of word remained robust, and retains the lowest concurvity of all predictors (for details, see the supplementary materials).  More in general, it also seems unlikely to us that differences in syntactic constructions can account for all word-specific pitch contours given that, compared to formal speech, informal conversations show much less variation in syntactic structure \citep{Tucker:Ernestus:2021}.


Finally, it is of course possible that the effect of word is driven by the segmental makeup of the words.  Since word and segmental makeup are confounded, this explanation is worth considering.  However, we have shown that pitch contours not only differ by word but also by (token-specific) meaning. We therefore ague that another factor is at play, a new kid on the block in phonetics: word meaning as operationalized with distributional semantics.

When word meanings are understood as abstract symbols that are associated with lower-level abstract units such as phonological words and their associated phonemes and tones \citep[as in, e.g., the WEAVER model of speech production,][]{levelt1999theory},  then the role that a word's meaning has is limited to selecting the proper associated phonological word. 
In addition to a word's phonological specification, its exact phonetic realization is then further co-determined by many other factors such as speech rate, speaker design \citep{lindblom1990explaining}, normalized segment duration \citet{seyfarth2014word}, utterance position \citep{klatt1976linguistic,turk2007multiple}, word category \citep{lohmann2018cut}, morphosyntactic function \citep{plag2015homophony,loo2023paradigmatic}, information load \citep{Bell:et:al:2003, Aylett:Turk:2004,VanSon:Pols:2003}, the many social factors studied in sociolinguistics \citep[see, e.g.,][]{hay2010stuffed},  and effects of alcohol \citep{pisoni1989effects}.

Homophones provide an especially clear window on the limitations of conceptualizing word meanings in terms of abstract symbols.  Early work on English homophones \citep{gahl2008time,lohmann2018cut} clarified that homophones have spoken word durations that differ in the mean depending on their frequency of use, with higher-frequency homophones (e.g. \textit{time}) on average having shorter durations that their low-frequency counterparts (e.g., \textit{thyme}).  More recently, \citet{gahl_time_2024} have shown, using embeddings from distributional semantics, that the spoken word duration of English heterographic homophones is also co-determined by the semantic similarity of the homophones and the amount of support that homophone's segments receive from their embeddings.  

The results we report here takes the approach of \citet{gahl_time_2024} to a new explanatory variable: the F0 contours of Mandarin words.  Thanks to the availability of contextualized embeddings, it is possible to evaluate the consequences for phonetic realization of much more fine-grained differences is meaning than is possible with semantic symbols (which by themselves allow only same-different assessments) and their associated segments and tones. 
The results obtained in the present study, along with those reported by \citet{Chuang:Bell:Tseng:Baayen:2024} and \citet{lu2025realization,lu2024form}, show that contextualized embeddings are a novel tool for understanding the fine details of the phonetic realization of Mandarin tone \citep[see also][]{stein2021morpho,schmitz2021durational}.   Distributional semantics is a new kid on the block that makes it possible to explore and chart the surprisingly good alignments between semantic detail in context, and phonetic realization in context.

\begin{figure}
    \centering
    \includegraphics[width=1\linewidth]{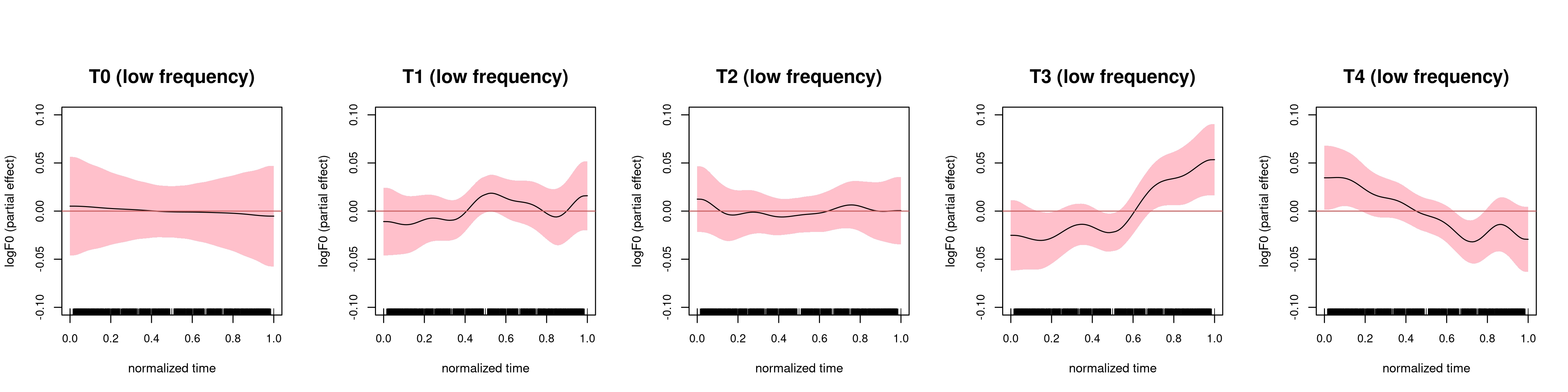}
    \caption{The partial effects for \texttt{tone pattern} in a GAM model fitted to low-frequency words.}
    \label{fig:freq}
\end{figure}

Our finding that words have their own --- semantically --- motivated pitch contours is yoked with further findings that provide additional challenges for the standard theory of Mandarin tone. First, our study only partially supports the canonically assumed contours of the four tones for Taiwan Mandarin.  Our study supports that  T4 is a falling pitch, but T2 and T3 appear to be level pitch contours, which was also found in \cite{fon1999does, hsu1999language}, just as T1, the only difference between these three tones being that T1 has higher pitch than T2 and T3 (see Figure~\ref{fig:tone pattern}).  
It might be argued that the flat tonal contours observed for T2 and T3 are due to tonal reduction, given that the words that we included for analysis have high token frequencies. High frequency words tend to undergo phonetic reduction in spontaneous speech \citep{Johnson:2004,Ernestus:Baayen:Schreuder:2002,Aylett:Turk:2004}. As a consequence, it is conceivable that lower-frequency words are more likely to be realized with their canonical tones as described by, e.g., \citep{xu1997contextual,peng1997production}. We therefore collected the tokens of all  monosyllabic word types that occurred at most  three times in the corpus, and preprocessed these tokens in the same way as described in  section~\ref{sec:data}. This resulted in an additional set of 286 word tokens representing 186 word types.  Figure~\ref{fig:freq} presents the partial effect of canonical tone for this set of low-frequency words, using the baseline GAM model specification enriched with \textbf{canonical tone} as predictor. (For the corresponding partial effects of tone for the main dataset, see Figure~\ref{fig:tone pattern}.) No differences in pitch height were present for the five tones as realized on low-frequency words. 
Furthermore, there were no significant contours for T0, T1 and T2, just as in our original data set. In contrast, T4 has a significantly falling pitch contour ($p < 0.0001$), also as in our original data set, and T3 has a late rise ($p < 0.0001$), which we did not find before.  This analysis does not support the hypothesis that canonical tones are visible for low-frequency words. It is possible that, due to the unavoidably small number of tokens of the low-frequency words, we may not have the statistical power to detect the presence of the canonical tones. What we do see is that the lowest frequency words have tone patterns that are very similar to those of the higher-frequency words.  This makes it highly unlikely that tonal reduction is at issue.  
Tonal reduction, furthermore, would predict the absence of word/meaning-specific pitch contours, which, unlike the canonical tone patterns, emerge loud and clear from our statistical analyses. We conclude that the descriptions of Mandarin tones in text books does not reflect their realization on monosyllabic words in conversational Taiwan Mandarin.
It may be correct for careful speech \citep{lai2008mandarin}, although we anticipate that word-specific tonal realizations can also be observed here.  For disyllabic words, canonical tone patterns appear better supported, albeit with small effect sizes \citep{lu2025realization}. \footnote{
Although the neutral tone is described in the literature on Beijing Mandarin as being dependent for its realization on the preceding tone \citep{chao1930system,yip1980tonal,chien2021investigating}, but see \cite{chao1968grammar}, our investigations of Taiwan Mandarin suggest that the neutral tone is a low-to-mid rising tone that is subject to the same variable effects of surrounding tones as the lexical tones. Our results are thus in line with the conclusion reached by \citep{huang2012study} that in Taiwan Mandarin the neutral tone is a tone in its own right. However, while \citep{huang2012study} found that in lab speech the neutral tone is a mid to low tone, our analyses for spontaneous speech support a low-to-mid rise.}

Our finding that the pitch contours on monosyllabic words hardly reflect their canonical tone patterns diverges considerably from what the textbooks, and standard phonological theories of Mandarin tone, would lead one to expect. In Mandarin textbooks, descriptions of the tonal system of Mandarin Chinese take the canonical lexical tones as givens \citep{xinhuadictionary}. This may be correct for careful speech \citep{lai2008mandarin}, although we anticipate that word-specific tonal realizations can also be observed here.  For conversational Taiwan Mandarin, it is becoming clear that the canonical tones are mostly not there, at least for monosyllabic words. For disyllabic words, canonical tone patterns are better supported, albeit with small effect sizes \citep{lu2025realization}. Importantly, across monosyllabic and disyllabic words, robust effects of word/meaning specific tonal signatures are present. 

The divergence between the actual realization of tones on monosyllabic words in conversational Taiwan Mandarin and their descriptions in textbooks can be due to many different factors, including dialectal differences and differences in speech register.   The possibility that the canonical tones of monosyllabic words are predominantly a feature of the formal registers of (some variant of) Mandarin Chinese fits well with the observation that when Chinese characters are taught in school, learners are taught not only to memorize the relevant strokes, but also to memorize the lexical tones associated with these characters --- tones that can be different from those used in their own dialects and their own conversational speech.  If the canonical tones were a feature of conversational speech that are learned before the onset of literacy, there should not be any need for such explicit instruction.

We now turn to the question of why word and meaning-specific pitch contours exist. Standard phonological theory does not provide a clear answer to this question. The observation that word-specific effects are often larger than just noticeable differences, and hence can in principle be perceived,  in combination with the fact that a simple machine learning algorithm can predict pitch contours from contextualized embeddings with accuracies exceeding a permutation baseline, fits with standard insights concerning how knowledge is transmitted across generations --- word-specific pitch contours are learnable.

Whereas standard phonological theory does not provide an answer to the question of why word-specific tonal signatures exist, a theoretical framework that actually predicts the existence of word-specific tonal signatures is provided by the Discriminative Lexicon Model \citep{baayen2019discriminative,Heitmeier:Chuang:Baayen:2025}.  This model implements mappings between high-dimensional representations of meaning (word embeddings) and high-dimensional representations of form. In the simplest case, these mappings are linear, and mathematically identical to the kind of mapping that was used above to predict time-normalized pitch contours from contextualized embeddings.  Thus, the DLM is a computational model of the lexicon without representations for words, nor for exemplars. The DLM does not make use of underlying representations, nor of rules or constraints that govern how surface forms are created from underlying forms.  Experience with understanding and producing words accumulates in networks that map meanings onto forms in production, and forms onto meanings in comprehension. The linear mappings that we used to predict pitch contours from contextualized embeddings thus fit perfectly within the DLM approach. In the present study, we have focused on the shape of pitch contours, for a study predicting spoken word duration using the DLM, see \citep{gahl_time_2024}.

\begin{figure}
    \centering
    \includegraphics[width=1\linewidth]{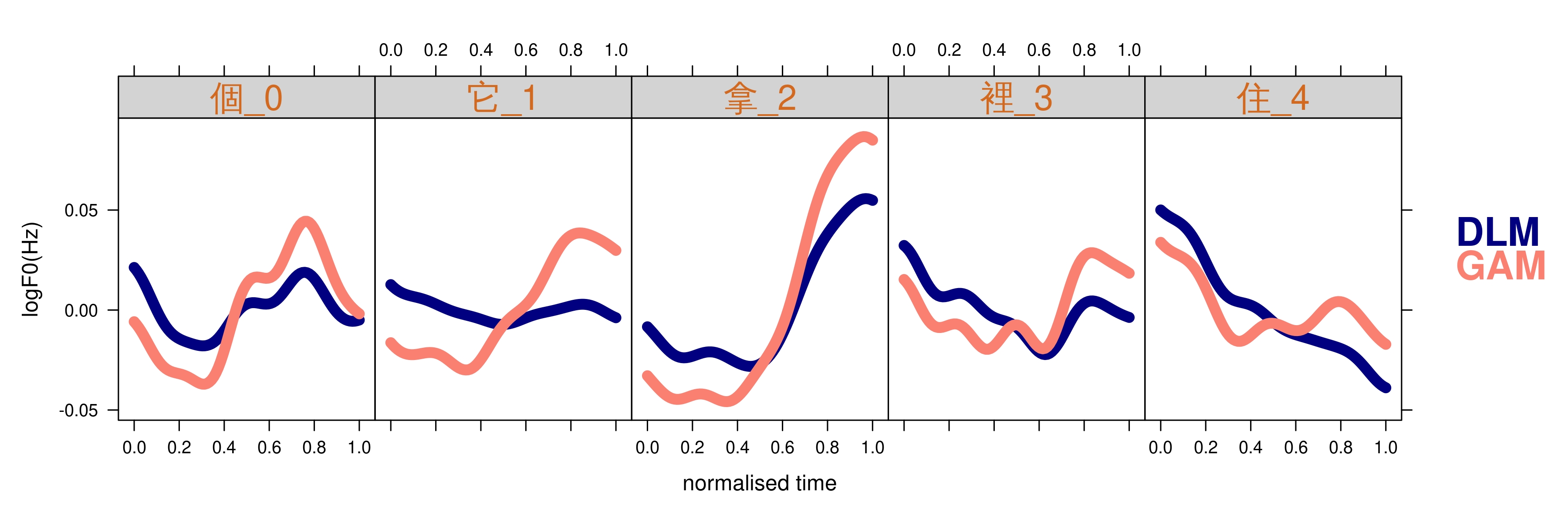}
    \caption{Pitch contours predicted from contextualized embeddings (blue) and the corresponding pitch contours identified by the GAMs (orange).}
    \label{fig: DLM_GAM}
\end{figure}

Interestingly,  as shown by \citet{Chuang:Bell:Tseng:Baayen:2024}, the word-specific pitch contour that the GAM reveals for a given disyllabic word with the T2-T4 tone pattern is very similar to the pitch contour that the DLM model predicts when given the centroid of the contextualized embeddings of the tokens of that word.  Furthermore, \citet{lu2025realization} shows that the tonal signature identified by the GAM for a given tone pattern (e.g., T2-T4, or T4-T4, or T3-T0), is very similar to the tonal contour predicted by the DLM when given as input the centroid of all tokens of all words sharing the same tone pattern.  Thus, the word-specific tone signatures, as identified by the GAM,  are best interpreted as the prototypical properties of forms (the mean pitch contours isolated by the GAM from the pitch contours of tokens) that are aligned with  prototypical meanings (the centroids of the contextualized embeddings of the corresponding tokens).  Figure~\ref{fig: DLM_GAM} presents the pitch contours identified by the GAM (orange) for 5 word types and the pitch contours predicted from the centroids of the contextualized embeddings of the tokens of these word types (blue). With the exception of 它, the similarities of predicted and observed pitch contours are striking.

In conclusion, our study offers a comprehensive exploration of the realization of the pitch contours of monosyllabic words in a corpus of spontaneous Taiwan Mandarin. We leveraged the analytical power of the generalized additive model, word embeddings, and the Discriminative Lexicon Model to understand the way in which contextualized embeddings, a new kid on the block, contribute to shaping F0 contours. We showed that, once word meanings are taken into account using the precision afforded by distributional semantics,  the realization of pitch in Mandarin can be predicted with greater precision than with standard approaches that take the canonical tones as givens.  We suspect that with our study we have just scratched the surface of the intricacies of Mandarin tone and of the role of word meaning in phonetics.

\clearpage
\appendix
\section*{Appendix}
\setcounter{figure}{0}

\section{Word senses distinguished for 16 characters}\label{appendix translation}
\setcounter{table}{0}
\renewcommand{\thefigure}{A\arabic{figure}}
Complementing Figure~\ref{fig:sense}, the following table lists the word senses and their definitions in Mandarin and English.

\begin{CJK}{UTF8}{bsmi}
\vspace*{2\baselineskip}
\nopagebreak
\footnotesize{
\begin{longtable}{p{3cm}p{6cm}p{8cm}}
\hline 

          \textbf{Character\_Word sense} & \textbf{Sense Meaning} & \textbf{Translation}\\\hline
          了\_sense\_1 &  表事件的結束或完整性 & indicate the completion or entirety of an event\\
          了\_sense\_3 &  清楚地知道 & clearly indicate awareness or realization \\
          了\_sense\_4 &  表前述對象對該事件或動作有實現的能力 & indicate that the subject of the previous clause has the ability to carry out the event or action\\
          
          個\_sense\_2 &  形容單獨一人 & describe a single individual \\
          個\_sense\_4 &  計算特定抽象名詞的單位 & a measure word for certain abstract nouns \\

          啊\_sense\_1 &  表解釋或提醒對方的語氣 & indicate an explanatory or reminding tone\\
          啊\_sense\_5 &  表停頓的語氣 & indicate a pausing tone\\
          啦\_sense\_1 &  表對前述評語內容具緩和作用的語氣 & indicate a moderating tone towards the aforementioned evaluative content\\
          啦\_sense\_3 &  表停頓的語氣 & indicate a pausing tone\\
          啦\_sense\_4 &  表驚訝的詢問語氣 & indicate a surprised questioning tone\\
          嗎\_sense\_1 &  表停頓的語氣 & indicate a pausing tone\\
          嗎\_sense\_2 &  表對前述命題疑問的語氣 & indicates a question regarding the previous proposition\\
          嗎\_sense\_3 &  表徵詢對方意願的疑問語氣 & indicate an inquiry into the other party’s intention\\
      
          大\_sense\_1 &  形容程度超過比較對象的 &  something  is greater than the object being compared\\
          大\_sense\_2 & 形容體積超過比較對象的 & bigger in capacity\\
          大\_sense\_3 &  形容年紀比特定對象大 & elder/older than somebody\\
          大\_sense\_4 &  形容年紀大的 & older age\\
          
          把\_sense\_2 &  計算用手動作次數的單位，前接數詞，多半出現在動詞的後面 & a measure word for counting hand movements, usually appears after verbs and preceded by numerals\\
          把\_sense\_5 &  握而使用 & using hand to take or hold things\\
          把\_sense\_6 &  引介導致狀態改變的目標 & to introduce a target that causes a state change\\

          的\_sense\_1 &  表肯定或加強的語氣 & indicate affirmation or emphasis\\
          的\_sense\_2 &  列舉相類似的事物 & list similar items\\
          的\_sense\_3 & 前述對象連接的是後述對象，表兩者指涉相同 & indicate that the previous and following object refer to the same\\
          的\_sense\_5 &  表以前述動作的狀態 & indicate the state resulting from a previous action\\
          路\_sense\_2 &  比喻能達到特定目標的途徑 & metaphorically refers to a pathway to achieve a specific goal\\
          路\_sense\_3 &  從起點到終點的距離 & the distance from the starting point to the ending point\\
         
          部\_sense\_1 &  計算編寫而成的作品的單位，包括書籍、戲劇、音樂等 & a measure word for works such as books, dramas, music, etc\\
          部\_sense\_2 &  整體中可區分的組成單位 & a distinguishable component unit within a whole\\\hline       
    \hline
    \caption{Word senses referenced in Figure~\ref{fig:sense}.}
    \label{tab:sense_translation}
\end{longtable}
}
\end{CJK}

\newpage

\bibliography{references}

\end{CJK}
\end{CJK}
\end{document}